\documentclass[pdflatex,sn-mathphys-num]{sn-jnl}%

\usepackage{graphicx}
\usepackage{caption}
\usepackage{subcaption}
\usepackage{booktabs}
\usepackage{longtable}

\usepackage{dsfont}
\usepackage{graphicx}%
\usepackage{multirow}%
\usepackage{amsmath,amssymb,amsfonts}%
\usepackage{amsthm}%
\usepackage{mathrsfs}%
\usepackage[title]{appendix}%
\usepackage{xcolor}%
\usepackage{textcomp}%
\usepackage{manyfoot}%
\usepackage{booktabs}%
\usepackage{algorithm}%
\usepackage{algorithmicx}%
\usepackage{algpseudocode}%
\usepackage{listings}%
\usepackage[most]{tcolorbox}
\usepackage{adjustbox}
\usepackage{array}
\usepackage[utf8]{inputenc}
\usepackage{threeparttable}

\usepackage{soul} %
\sethlcolor{green!18} %

\usepackage{tikz}
\usetikzlibrary{arrows.meta,positioning}

\usepackage{fancyvrb}      %
\usepackage[T1]{fontenc}   %
\usepackage{lmodern}       %
\usepackage{amsmath}
\usepackage{cleveref}

\makeatletter

\def\Titlefont{\reset@font\fontsize{17bp}{22.5bp}\selectfont\raggedright}%
\def\SubTitlefont{\reset@font\fontsize{14bp}{16.5bp}\selectfont\raggedright}%
\def\Authorfont{\reset@font\fontsize{12bp}{14.5bp}\selectfont\boldmath\raggedright}%
\def\addressfont{\reset@font\fontsize{11bp}{13.5bp}\selectfont\raggedright}%
\def\abstractheadfont{\reset@font\fontsize{9bp}{11bp}\bfseries\selectfont\raggedright}%
\def\abstractfont{\reset@font\fontsize{9bp}{11bp}\selectfont\raggedright}%

\renewcommand\normalsize{%
   \@setfontsize\normalsize{9.5bp}{11.4bp}%
   \abovedisplayskip 12\p@ \@plus2\p@ \@minus1\p@
   \abovedisplayshortskip \z@ \@plus3\p@%
   \belowdisplayshortskip 3\p@ \@plus3\p@ \@minus3\p@%
   \belowdisplayskip \abovedisplayskip%
   \let\@listi\@listI}%

\renewcommand\small{%
   \@setfontsize\small{8.5bp}{10.2bp}%
   \abovedisplayskip 5\p@ \@plus3\p@ \@minus4\p@
   \abovedisplayshortskip \z@ \@plus2\p@
   \belowdisplayshortskip 3\p@ \@plus2\p@ \@minus2\p@
   \def\@listi{\leftmargin\leftmargini
               \topsep 4\p@ \@plus2\p@ \@minus2\p@
               \parsep 2\p@ \@plus\p@ \@minus\p@
               \itemsep \parsep}%
   \belowdisplayskip \abovedisplayskip}%

\renewcommand\footnotesize{%
   \@setfontsize\footnotesize{7bp}{8bp}%
   \abovedisplayskip 5\p@ \@plus2\p@ \@minus4\p@
   \abovedisplayshortskip \z@ \@plus\p@
   \belowdisplayshortskip 3\p@ \@plus\p@ \@minus2\p@
   \def\@listi{\leftmargin\leftmargini
               \topsep 3\p@ \@plus\p@ \@minus\p@
               \parsep 2\p@ \@plus\p@ \@minus\p@
               \itemsep \parsep}%
   \belowdisplayskip \abovedisplayskip}%

\def\sectionfont{\reset@font\fontfamily{\rmdefault}\fontsize{12bp}{14bp}\bfseries\selectfont\raggedright\boldmath}%
\def\subsectionfont{\reset@font\fontfamily{\rmdefault}\fontsize{10.5bp}{12.5bp}\bfseries\selectfont\raggedright\boldmath}%
\def\subsubsectionfont{\reset@font\fontsize{9.5bp}{11.5bp}\bfseries\selectfont\raggedright\boldmath}%
\def\paragraphfont{\reset@font\fontsize{9bp}{11bp}\bfseries\itshape\selectfont\raggedright}%

\renewcommand{\@maketitle}{\null%
    \if@remarkboxon\vbox to 0pt{\vspace*{-78pt}\hspace*{-18pt}\FMremark}\else\vskip-10pt\fi%
    \hsize\textwidth\parindent0pt%
    {\hbox to \textwidth{{\Artcatfont\ArtType\hfill}\par}}%
    \ifx\@title\empty\else%
        \removelastskip\vskip20pt\nointerlineskip%
        {\Titlefont\@title\par}%
    \fi%
    \ifx\@subtitle\empty\else%
        \vskip9pt%
        {{\SubTitlefont\@subtitle\par}}%
    \fi%
    \ifnum\aucount>0
        \global\punctcount\aucount%
        \vskip20pt%
        \artauthors\par%
        {\vskip7pt\addressfont\auaddress\par%
	 \removelastskip\vskip24pt%
	\ifnum\emailcnt>0\relax%
           \ifx\corrauthemail\@empty\else{\ifnum\aucount>1*\fi}%
	   Corresponding author(s). E-mail(s): \corrauthemail\par\fi%
	   \ifx\authemail\@empty\else Contributing authors:\ \authemail\fi%
        \fi%
        \ifequalcont{\par$^{\dagger}$\@equalconttext\par}\fi%
	 \removelastskip\vskip24pt%
        \ifpresentaddress{\par\@presentaddresstext\par}\fi%
	}%
     \fi%
     {\printabstract\par}%
     {\printkeywords\par}%
     \ifx\@pacs\empty\else%
       \loop\ifnum\PacsCount>0%
          \csname\romannumeral\PacsTmpCnt StorePacsTxt\endcsname\par%
          \StepDownCounter{\PacsCount}%
          \StepUpCounter{\PacsTmpCnt}%
       \repeat%
    \fi%
    \removelastskip\vskip36pt\vskip0pt}%

\makeatother

\theoremstyle{thmstyleone}%

\theoremstyle{thmstyletwo}%

\theoremstyle{thmstylethree}%

\begin{document}

\title[Article Title]{Temporally annotated textual time series from PubMed Open Access clinical case reports}

\author[1,2]{\fnm{Shahriar} \sur{Noroozizadeh}}\email{snoroozi@cs.cmu.edu}
\equalcont{These authors contributed equally to this work.}

\author*[3]{\fnm{Sayantan} \sur{Kumar}}\email{sayantan.kumar@nih.gov}
\equalcont{These authors contributed equally to this work.}

\author[2,1]{\fnm{George H.} \sur{Chen}}\email{georgechen@cmu.edu}

\author[3]{\fnm{Jeremy C.} \sur{Weiss}}\email{jeremy.weiss@nih.gov}

\affil[1]{\orgdiv{Machine Learning Department, School of Computer Science}, \orgname{Carnegie Mellon University}, \city{Pittsburgh}, \state{Pennsylvania}, \country{USA}}

\affil[2]{\orgdiv{Heinz College of Information Systems and Public Policy}, \orgname{Carnegie Mellon University}, \city{Pittsburgh}, \state{Pennsylvania}, \country{USA}}

\affil[3]{\orgdiv{National Library of Medicine}, \orgname{National Institutes of Health}, \orgaddress{%
\city{Bethesda}, \state{Maryland}, \country{USA}}}

\abstract
{
Clinical narratives encode temporal dynamics essential for modeling patient trajectories, yet large-scale temporally annotated resources are scarce. We introduce PMOA-TTS, a corpus of 124,699 single-patient PubMed Open Access case reports converted into structured textual timelines of (event, time) pairs using a scalable large-language-model pipeline (\texttt{Llama 3.3 70B} and \texttt{DeepSeek-R1}). The corpus comprises over 5.6 million timestamped events, alongside extracted demographics and diagnoses. Technical validation uses a clinician-curated gold set and three measures: semantic event matching, temporal concordance (c-index), and alignment error summarized with Area Under the Log-Time CDF (AULTC). We benchmark alternative prompting and model choices and provide documentation to support reproduction. PMOA–TTS enables research on timeline extraction, temporal reasoning, survival modeling and event forecasting from narrative text, and offers broad diagnostic and demographic coverage. 
Data and code are openly available in public repositories.
}.

\keywords{clinical natural language processing, temporal information extraction, textual time series, large language models, PubMed Open Access case reports}

\maketitle

\section{Introduction}
\label{sec:introduction}

Understanding when clinical events occur is fundamental to modeling patient trajectories, enabling tasks such as process mining, outcome forecasting, and causal inference in medicine \citep{jensen2012mining, noren2010temporal}.
However, most clinical narratives lack explicit temporal structure. While electronic health record (EHR) datasets like MIMIC-III and MIMIC-IV include timestamped structured data and clinical notes \citep{johnson2016mimic, johnson2023mimic}, they capture only a subset of events (e.g., vital signs, lab results) and still require sophisticated natural language processing to extract temporal information from unstructured text \citep{viani2021temporal, cheng2023typed}.

In contrast, published case reports often provide rich narrative descriptions of patient timelines, but the timing of events is typically expressed in relative terms within the free text (e.g., “on day 3 of hospitalization”).
Efforts in temporal reasoning on clinical text have been hampered by the scarcity of large-scale annotated corpora.
Earlier initiatives, such as the 2012 i2b2 challenge and Clinical TempEval, focused on temporal relation extraction from clinical notes \citep{sun2013evaluating, bethard2016semeval, galvan2018investigating} but were constrained by small datasets (e.g., 310 documents) from single institutions, limiting model performance and generalizability.
Systems that rely on metadata timestamps (e.g., admission or discharge dates) fail to capture the fine-grained sequence of clinical events described in text, which is essential for nuanced understanding and prediction.
There remains a pressing need for large-scale, temporally annotated corpora to support the development and evaluation of models for clinical temporal reasoning.

Building on our pilot pipeline on 10 case reports \citep{wang2025large}, we introduce the PubMed Open Access Textual Time Series (PMOA–TTS) corpus—a large-scale dataset of 124,699 openly available clinical case reports, each annotated with a structured timeline of events.
Leveraging large language models (LLMs)—specifically \texttt{Llama 3.3 70B} \citep{grattafiori2024llama} and \texttt{DeepSeek-R1} \citep{guo2025deepseek}—we transform each case report into a sequence of \mbox{(clinical event, time)} tuples that capture key clinical events and their relative timing within the patient narrative (Figure \ref{fig:annotation_ex}).
To our knowledge, PMOA–TTS constitutes the largest publicly available collection of clinical narratives with explicit temporal event structure.
We demonstrate that these temporal annotations enable novel forms of analysis and significantly enhance downstream modeling of patient outcomes. 
Compared to our prior pilot study \cite{wang2025large}, we expand the manually annotated reference set from 10 to 200 case reports and scale the LLM-annotated timelines from 100 to 124,699 reports. To our knowledge, PMOA–TTS constitutes the largest publicly available collection of clinical narratives with explicit temporal event structure.

Building PMOA-TTS requires addressing three challenges at scale: reliably identifying single-patient case reports, extracting timeline events with high fidelity from ambiguous free-text expressions, and rigorously evaluating the resulting annotations. We meet these needs with an LLM-based pipeline that selects relevant narratives beyond
metadata filters, resolves relative time language into fine-grained event--time tuples, and enriches each case with structured demographics and diagnoses. To illustrate
downstream
value, we also formulate a
survival analysis task that operates directly on the extracted textual time series.

\begin{figure*}[t]
  \centering

  \begin{subfigure}[t]{0.49\textwidth}\vspace{0pt}
    \begin{tcolorbox}
    [
    colback=white,
    colframe=black!30,
    arc=2mm,
    boxrule=0.6pt,
    left=2mm,right=2mm,top=2mm,bottom=2mm
    ]
      \small\raggedright
      A \hl{52-year-old} \hl{woman} was \hl{admitted to the medical ICU} with \hl{chest pain} and \hl{shortness of breath}. She developed \hl{intermittent fever and fatigue} a couple of weeks prior to admission. Eight hours after hospital and ICU admission, \hl{broad-spectrum antibiotics} were \hl{started for suspected infection}. On hospital admission day 2, she developed \hl{pleuritic pain} and \hl{escalating oxygen requirements}. \hl{CT angiography} was performed two days after admission.
      
    \end{tcolorbox}
    \caption{Case narrative with highlighted clinical events}
    \label{fig:narrative_box}
  \end{subfigure}\hfill%
  \begin{subfigure}[t]{0.49\textwidth}\vspace{0pt}
    \begin{tcolorbox}
    [
    colback=white,
    colframe=black!30,
    arc=2mm,
    boxrule=0.6pt,
    left=2mm,right=2mm,top=2mm,bottom=2mm
    ]
      \small\centering
      \setlength{\tabcolsep}{5pt}
      \begin{tabular}{@{} l r @{}}
        \toprule
        \textbf{Event} & \textbf{Time (h)} \\
        \midrule
        52-year-old & 0 \\
        woman & 0 \\
        admitted to the medical ICU & 0 \\
        chest pain & 0 \\
        shortness of breath & 0 \\
        intermittent fever & -336 \\
        fatigue & -336 \\
        broad-spectrum antibiotics started \\ for suspected infection & 8 \\
        pleuritic pain developed & 24 \\
        escalating oxygen requirements & 24 \\
        CT angiography & 48 \\
        \bottomrule
      \end{tabular}
    \end{tcolorbox}
    \caption{Text-ordered event–time tuples.}
    \label{fig:table_box}
  \end{subfigure}

  \caption{Example case report (left) with text-ordered event-time tuples (right).
  }
  \label{fig:annotation_ex}
  \vspace{-5mm}
\end{figure*}

In summary, we (i) introduce the PMOA Textual Time Series (PMOA-TTS) corpus of 124{,}699 single-patient case reports from PubMed Open Access, each annotated with structured clinical timelines extracted using large language models; (ii) develop a scalable pipeline for automatic case identification and prompt-driven extraction of timelines, demographics, and diagnoses, together with a manual evaluation framework; (iii) define and apply complementary validation measures for timeline accuracy, including temporal order concordance (c-index) and the Area Under the Log-Time CDF (AULTC); (iv) characterize corpus properties (demographics, frequent diagnoses, and co-occurrence structure) to document clinical and demographic breadth; and (v) illustrate one reuse pathway by framing survival prediction directly over textual time series.

PMOA–TTS offers several
contributions to the biomedical NLP and clinical AI communities. By releasing the largest openly available dataset of temporally structured clinical narratives, this work democratizes access to timeline-based clinical text—an area typically limited to proprietary EHR datasets. The corpus spans a wide range of diagnoses and demographics, supporting research on temporal patterns in diverse and sometimes underrepresented conditions.
By making the extraction pipeline, prompting strategies, and evaluation framework publicly available, the project promotes transparency and reproducibility in LLM-based clinical information extraction. These contributions collectively advance the development of models for timeline reconstruction, temporal reasoning, and longitudinal prediction, while lowering the barrier
to entry for researchers working in resource-limited settings.

\section{Methods}
\label{sec:methods}

\begin{figure}[!htbp] %
    \centering
    \includegraphics[width=0.75\linewidth]{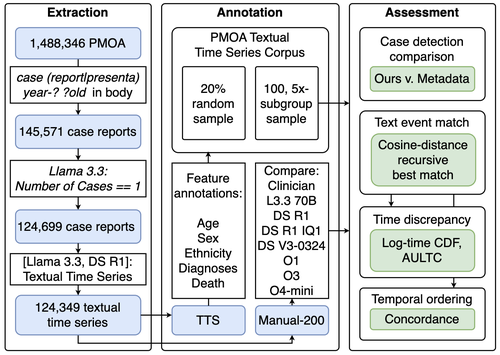} %
    \caption{Flowchart of the extraction and annotation pipeline. The left panel (Extraction) shows how we filtered the PMOA corpus to identify single-patient case reports. The middle panel (Annotation) depicts the generation of textual time series for each case via LLM prompting and the creation of evaluation subsets. The right panel (Assessment) summarizes the evaluation process, including metadata comparison, text event matching, time discrepancy analysis (log-time CDF, AULTC), and temporal order concordance.
    }
    \label{fig:flowchart}
    \vspace{-2.5mm}
\end{figure}

Our pipeline comprises the following key components: dataset extraction, textual time series annotation, and comprehensive evaluation of the annotations (Figure~\ref{fig:flowchart}).

\subsection{Case Report Identification Pipeline}
\label{sec:case-report-identification-pipeline}

We use the PubMed Open Access (PMOA) repository data, which includes 1.4 million published manuscripts. Specifically, we utilize the plain text files from the December 17, 2024 release, located in the \texttt{oa\_noncomm} folder, which supports both noncommercial and commercial use. Following our previous work in \cite{wang2025large}, we identify candidate case reports by extracting content between the section markers ``\texttt{==== Body}'' and ``\texttt{==== Ref}'', which separate the main text and references in the PMOA format.

To identify potential case reports, we filtered documents where the free-text match the case-insensitive regular expressions \textcolor{teal}{\texttt{case (report|presenta)}} and \textcolor{teal}{\texttt{year-? ?old}}.
This filter yields 145,571 candidate documents likely to be case reports or case series. To narrow this set to single-patient case reports, we use a large language model (\texttt{Llama 3.3 70B Instruct}) to analyze each document's body text. The model is prompted with: ``\texttt{You are a physician. Determine the number of case reports in the following manuscript. Return 0 if it is not a case report. Reply only with the number and nothing else.}'', followed by the document body. We retain documents for which the model replies with \texttt{1} and discard others. This two-stage process results in a final dataset of 124,699 single-patient case reports (Figure~\ref{fig:flowchart}).

We compared our LLM-based case report identification method with PubMed's metadata-based labeling using a carefully designed benchmark. We created a dataset of 100 reports across five diagnoses—Diabetes, Sepsis, Hypertension, COVID-19, and Atrial Fibrillation—with 10 reports per diagnosis from each method. A clinician manually reviewed and labeled each report to verify whether it described exactly one case.

\subsection{Timeline Annotation via LLM Prompting}
\label{sec:timeline-annotation}

\subsubsection{Textual Time Series (TTS)}
\label{sec:textual-time-series}

We use large language models (LLMs; \texttt{LLaMA~3.3~70B}, \texttt{DeepSeek-R1}) to convert PubMed Open Access (PMOA) case reports into \emph{textual time series} (TTS). A textual time series is defined as an ordered sequence of clinical event–time pairs, where each event is represented as a short free-text span and each timestamp is a numeric value measured in hours relative to a case-specific reference time. The reference time is defined as the time of hospital admission when an admission is explicitly described; if no admission time is stated, the earliest documented clinical encounter or presentation in the report is used as the reference point. Events occurring before the reference time are assigned negative timestamps, and events occurring after are assigned positive timestamps.

In this work, a clinical event refers to any health-related mention that can stand alone semantically and is directly relevant to the patient’s course. This includes symptoms and signs, diagnoses, procedures and diagnostic tests, treatments and medication administrations, major clinical states, and outcomes. Explicitly stated pertinent negatives and termination events (such as “no shortness of breath” or “antibiotics discontinued”) are also included. 
Demographic attributes that appear in the narrative, such as age and sex are also extracted as clinical events with reference time (t=0) as the timestamp. We exclude purely contextual text that does not describe the index patient, such as general background statements or literature discussion.

To preserve clinical nuance, we retain minimally edited, context-rich text spans rather than restricting events to short phrases. In contrast to span-based temporal annotation methods (e.g., \cite{sun2013evaluating, uzuner20112010}) and i2b2-style guidelines that often limit events to short prepositional phrases, our approach allows events to include sufficient surrounding context to accurately convey meaning (for example, retaining “pain in chest that radiates substernally” as a single event). Conjunctive mentions are separated into individual events when doing so improves clarity, and semantic modifiers that alter meaning—such as negation, uncertainty, or intent (e.g., “planned”)—are preserved in the event string so that intended actions are not conflated with actions that occurred. 

Natural-language time expressions are normalized into hour units: coarse expressions such as “hospital day 2” are converted using 24-hour increments from the reference point, interval expressions are represented using the start of the interval, and vague temporal phrases are resolved to approximate offsets that are consistent with narrative ordering and surrounding cues. The goal of this normalization is to produce a usable relative timeline rather than an exact reconstruction of clock time.

The above specifications define the intended representation of textual time series. The LLM extraction prompt used in our pipeline (Supplementary Material, Section~\ref{apd:extraction-prompt}) implements this specification by instructing the model to output a structured sequence of \mbox{(clinical event, time)} tuples. Events are kept in free-text form to retain clinical specificity rather than being normalized to a fixed ontology. On average, each case report yields 46 events, %
and the full corpus comprises over 5.6 million timestamped events, making it the largest publicly available resource of this type in clinical NLP.

\subsubsection{Gold Standard Annotations}
\label{sec:gold-standard-annotations}

We constructed a gold-standard reference set of 200 manually annotated case reports, consisting of 190 cases randomly sampled from PMOA–TTS and 10 cases from our prior pilot study. All annotations were performed by a single clinical expert following a detailed annotation guideline designed to preserve fidelity to the original narrative while producing event representations suitable for temporal evaluation.

Annotators marked exact text spans corresponding to clinical events without rewriting or paraphrasing, with a small number of controlled exceptions specified in the guideline. First, conjunctive findings (e.g., “nausea and vomiting”) were permitted to be split into individual events when clinically meaningful. Second, when needed to supply context, annotators could prepend short clarifying phrases such as “history of” to ensure that extracted events were interpretable as stand-alone findings. Beyond these two exceptions, events were expected to remain faithful to the wording and granularity of the source text.

To promote consistency, annotators were then instructed to ensure that each event was semantically self-contained (e.g., using “SCC of the lung” rather than “SCC” to avoid ambiguity with other
entities such as cutaneous squamous cell carcinoma or sickle cell crisis). Related findings were disambiguated when necessary, duplicate mentions of the same event were avoided even when they appeared in multiple sections of a report, and pertinent negative or termination events (e.g., “no shortness of breath” or “denies chest pain”) were included when explicitly stated. These conventions were applied uniformly across all 200 annotated case reports.

Each annotated event was assigned a timestamp relative to a standardized case reference point. By definition, this reference point was the time of admission. If an explicit admission time was not stated, the encounter date (the earliest unambiguous point of primary clinical evaluation within the case narrative) was used as the reference time. For temporal expressions that described intervals, the start time of the interval was recorded. All timestamps were expressed in hours relative to the reference time (t=0).

These manually produced timelines serve as the reference standard for evaluating LLM-generated annotations in subsequent sections.

\subsection{Evaluation Metrics}
\label{sec:evaluation-metrics}

We next assessed the quality of the extracted textual time series using the 200 manually annotated case reports as reference. Each large language model (LLM) annotator was evaluated across five independent runs to capture run-to-run variation in the generated timelines. For each model and run, we computed three complementary metrics as described below (detailed description in Supplementary Material, Section \ref{apd:evaluation-metrics}).

\paragraph{Event match rate}

\textbf{\emph{Event match rate}} measures how well the extracted clinical events match up with manually annotated reference events. To evaluate alignment between predicted and reference clinical events, we applied a recursive best-match strategy adapted from prior work \citep{wang2025large}, as summarized in Algorithm \ref{alg:recursive_match}. At each iteration, the method selects the closest unmatched predicted–reference pair under a text-similarity metric, retains the match if it falls below a distance threshold, and removes both events before continuing. This procedure yields an efficient one-to-one alignment that accommodates timelines of unequal length.

As a sensitivity analysis, we also evaluated a globally optimal one-to-one alignment based on the Linear Assignment Problem (LAPJV) \citep{jonker1987shortest}. Across all models, results differed by at most $0.3$ percentage points in event match rate relative to the recursive heuristic. We therefore report results using the recursive matching strategy in the main text and provide full LAPJV details and comparison tables in the Supplementary Material (Section~\ref{apd:lapjv-matching}).

\paragraph{Temporal metrics}

We used 2 metrics which evaluate how well the LLM extracts timestamps of clinical events: (a) \textbf{\emph{Temporal concordance}} (c-index) evaluates whether the relative ordering of event times in the model output agrees with the manually annotated timelines, providing a order-based summary of temporal consistency. (b) \textbf{\emph{Log-time discrepancy}} quantifies the absolute difference between the log-transformed timestamps of matched events, and we summarized its distribution using the Area Under the Log-Time CDF (AULTC), which captures the cumulative proportion of events achieving small timing errors.

\section{Data Records}
\label{sec:data-records}

The PMOA–TTS dataset is hosted on Hugging Face Datasets at \url{https://huggingface.co/datasets/snoroozi/pmoa-tts} and is released under \textbf{CC BY-NC-SA 4.0}. The dataset card lists task tags (\texttt{Text Classification, Time Series Forecasting}), modality (\texttt{Text}), format support (\texttt{Parquet via auto-conversion}), language (\texttt{English}), and size (“\texttt{100K–1M}”). Data record schema with an example JSON entry is shown in Figure~\ref{fig:record_schema}. PMOA-TTS has the following dataset splits: 

\begin{figure*}[t]
\centering

\begin{subfigure}[!]{1\textwidth}
\centering
\footnotesize
\begin{tabular}{@{} l l p{4cm} @{}}
\toprule
\textbf{Field} & \textbf{Type} & \textbf{Description} \\
\midrule
\texttt{pmc\_id} & str &
Folder-level ID (PMC000xxxxxx--PMC011xxxxxx). \\

\texttt{case\_report\_id} & str &
Filename of the case (e.g., PMC6417290). \\

\texttt{textual\_timeseries} & List[Dict] &
Sequence of \{"event": str, "time": int\}. \\

\texttt{demographics} & Dict &
\{"age": str/"Not Specified", "sex": str, "ethnicity": str\}. \\

\texttt{diagnoses} & List[str] &
List of diagnosis terms extracted per case. \\

\texttt{death\_info} & Dict &
\{"observed\_time": int, "death\_event\_indicator": 0/1\}. \\
\bottomrule
\end{tabular}
\end{subfigure}
\hfill

\begin{subfigure}[t]{0.65\textwidth}
\centering
\begin{tcolorbox}[
  colback=gray!5,
  colframe=gray!70,
  title=Example JSON entry,
  boxrule=1pt,
  enhanced,
  fontupper=\footnotesize
]
\begin{verbatim}
"pmc_id": "PMC006xxxxxx",
"case_report_id": "PMC6417290",

"textual_timeseries": [
  {"event": "56 years old", "time": 0},
  {"event": "male", "time": 0},
  {"event": "HIV-positive", "time": 0},
  {"event": "admitted to the hospital", "time": 0},
  {"event": "knee arthralgia", "time": 0},
  ...
  {"event": "postural headache", "time": 48},
  {"event": "neurological examination", "time": 120},
  {"event": "persistence of headache", "time": 144},
  {"event": "brain computed tomography scan",
   "time": 144},
  ...
  {"event": "symptom free", "time": 4320},
  {"event": "CT scan", "time": 4320},
  {"event": "no brain shift", "time": 4320}
],
"diagnoses": [
  "HIV",
  "Knee arthralgia",
  "Subdural hematoma",
  "Postdural puncture headache (PDPH)"
],
"death_info": {
  "observed_time": 4230,
  "death_event_indicator": 0
},
"demographics": {
  "age": 56,
  "sex": "Male",
  "ethnicity": "Not Specified"
}
\end{verbatim}
\end{tcolorbox}
\end{subfigure}

\caption{\textbf{top:} PMOA-TTS record schema. \textbf{bottom:} compact example matching the schema above (drawn from our HuggingFace repository (formatted for readability).}
\label{fig:record_schema}
\vspace{-3mm}
\end{figure*}

\begin{itemize}
    \item \texttt{train\_DSR1/L33}: The full dataset of 125k single-patient case reports automatically extracted using \texttt{DeepSeek-R1} and \texttt{Llama 3.3 70B} respectively. Here, a split denotes a version of the dataset defined by the annotation model used. These splits serve as the default releases and contain the primary annotations used in downstream analyses. %
    
    \item \texttt{case\_study\_100}: A curated benchmark of 100 case reports used to evaluate LLM-based versus metadata-based case report identification. Each report has been manually reviewed for single-case validity, meaning that the narrative describes the clinical course of exactly one individual patient (i.e., no aggregated cohorts, comparative cases, or multi-patient summaries), across five diagnostic categories. %
    
    \item \texttt{case\_study\_25k\_L33}: A randomly sampled subset of 25,000 case reports from the full corpus, annotated using \texttt{LLaMA~3.3~70B}. This split (i.e., subset of cases) is used for downstream survival analysis experiments, providing a computationally manageable cohort for modeling time-to-event outcomes. %
    
    \item \texttt{case\_study\_25k\_DSR1}: A 25k-case subset of \texttt{DeepSeek-R1}-annotated reports (same case reports corresponding to \texttt{case\_study\_25k\_L33}) used for downstream survival analysis experiments. The \texttt{textual\_timeseries} and \texttt{death\_info} fields in this split are reconstructed from \texttt{DeepSeek-R1} outputs, enabling comparative analysis between models across an identical patient cohort.
    
    \item \texttt{case\_study\_n200}: A benchmark of 200 manually reviewed case reports used for qualitative and quantitative evaluation of LLM-generated annotations. Each model is evaluated across five independent runs on this subset to assess run-to-run variability in annotation consistency.

\end{itemize}

The data repository also has an additional data split \texttt{TTS\_n200\_additional\_models}, which bundles the single-expert manual timelines (the reference set) together with parallel LLM-generated textual time series for additional models, including open-weight models such as \texttt{DeepSeek-V3}, \texttt{DeepSeek-R1 quantized}, and \texttt{DeepSeek-V3-0324}, and proprietary OpenAI models such as \texttt{GPT5}, \texttt{O1}, \texttt{O3}, and \texttt{O4-mini}. This split is designed for lightweight benchmarking and cross-model comparison on a fixed 200-case cohort without re-annotating the full corpus. For the full 125k release, we provide textual time series produced by \texttt{DeepSeek-R1} and \texttt{Llama 3.3 70B}; the additional models are released only on the 200-case subset due to resource constraints. For completeness, both \texttt{Llama 3.3 70B} and \texttt{DeepSeek-R1} were also executed in five independent runs on the \texttt{case\_study\_n200} subset, and these annotated outputs are included in the dataset.

\section{Data Overview}
\label{sec:data-overview}

\subsection{Cohort Demographics}
\label{sec:cohort-demographics}

For demographic extraction, we used an LLM-based query to retrieve age, sex, and ethnicity for each case. 
The prompt instructed the model to report age in years; sex as \texttt{Male}, \texttt{Female}, a free-text category, or \texttt{Not Specified}; and ethnicity according to U.S.\ Census categories or \texttt{Not Specified} (full prompt in Supplementary Material, Section~\ref{apd:demographics-prompt}).

\begin{figure}[!htbp]
  \centering
  \begin{minipage}[t]{0.49\linewidth}\vspace{0pt}
    \centering
    \includegraphics[width=\linewidth]{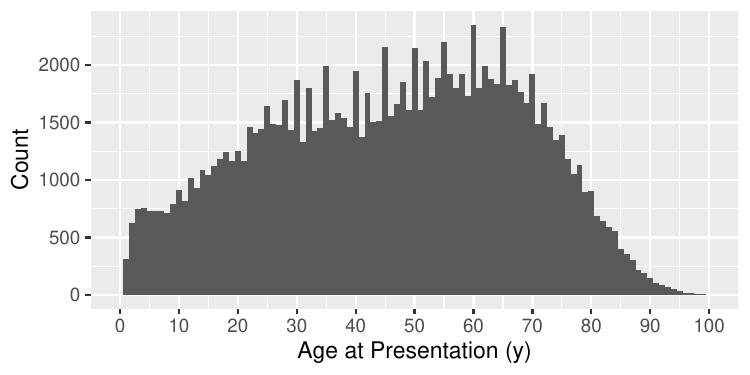}
  \end{minipage}\hfill
  \begin{minipage}[t]{0.49\linewidth}\vspace{0pt}
    \centering
    \scriptsize
    \begin{tabular}{l r}
      \toprule
      \textbf{Characteristic} & \textbf{Percent (\%)}\\
      \midrule
      \multicolumn{2}{l}{\textbf{Sex}}\\
      \quad Male & 52\\
      \quad Female & 47\\
      \quad Not Specified & 1\\
      \midrule
      \multicolumn{2}{l}{\textbf{Ethnicity}}\\
      \quad White/Caucasian & 5\\
      \quad Asian & 3\\
      \quad African & 1\\
      \quad Other & 2\\
      \quad Not Specified & 89\\
      \bottomrule
    \end{tabular}
  \end{minipage}
  \caption{Age, sex and ethnicity of 124,699 PMOA case reports}
  \label{fig:demo}
  \vspace{-7.5mm}
\end{figure}

Across PMOA-TTS, the median age is 47 years, with peaks at 5-year intervals, consistent with rounded age reporting in clinical text (Figure~\ref{fig:demo}). 
The inferred sex distribution is approximately balanced (52\% male, 47\% female, 1\% not specified). 
Ethnicity is rarely documented: 89\% of case reports lack explicit ethnicity, and among those that do, White/Caucasian (5\%) and Asian (3\%) are most common, followed by African (1\%) and Other (2\%). 
The relative enrichment of Asian patients likely reflects the international authorship of PubMed case reports rather than any single healthcare system.

\subsection{Descriptive Statistics}
\label{sec:descriptive-statistics}

\begin{figure}[!htbp]
    \centering
    \begin{minipage}{0.48\textwidth}
        \centering
        \includegraphics[width=\linewidth]{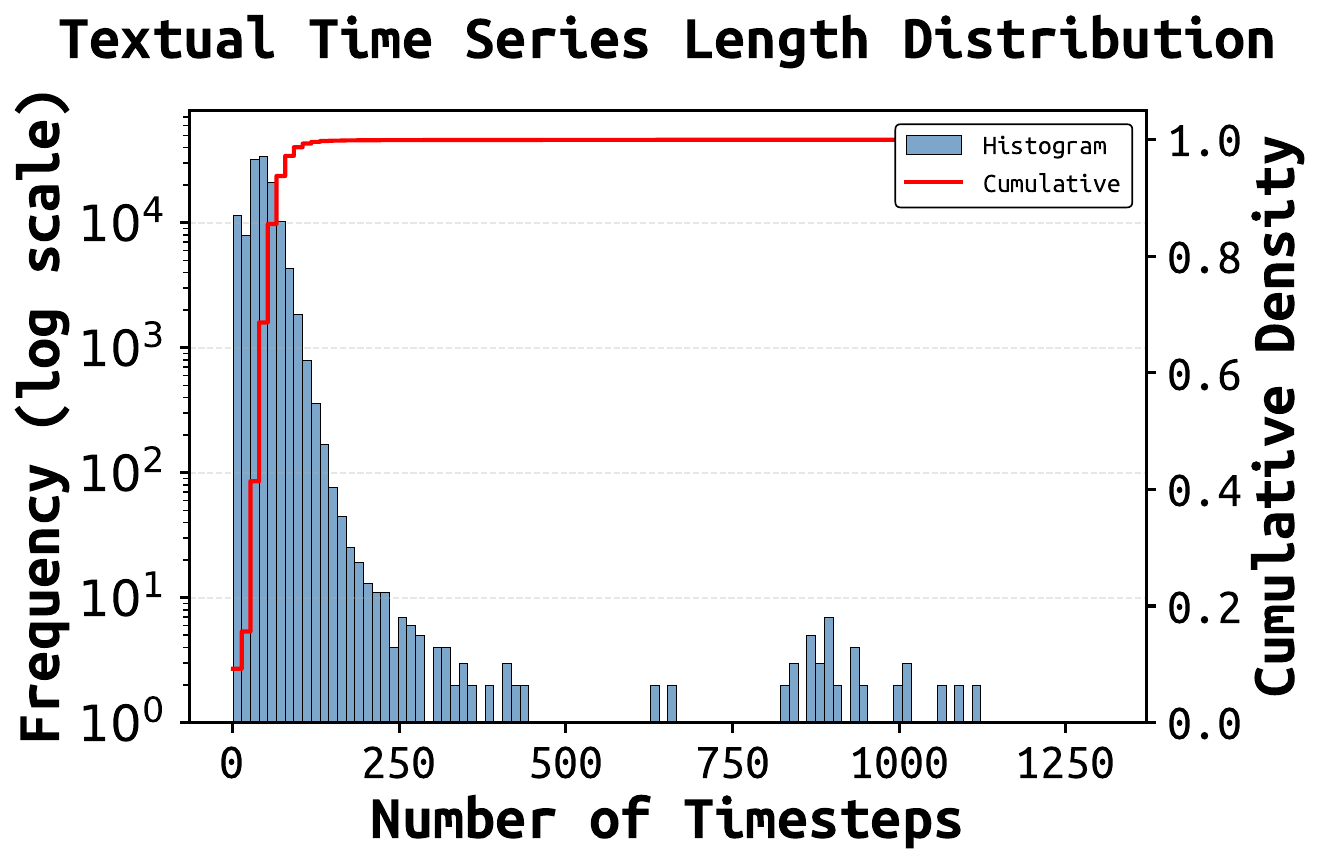}
        \caption*{(a) Distribution of time series lengths (timesteps) across the dataset.}
    \end{minipage}\hfill
    \begin{minipage}{0.48\textwidth}
        \centering
        \includegraphics[width=\linewidth]{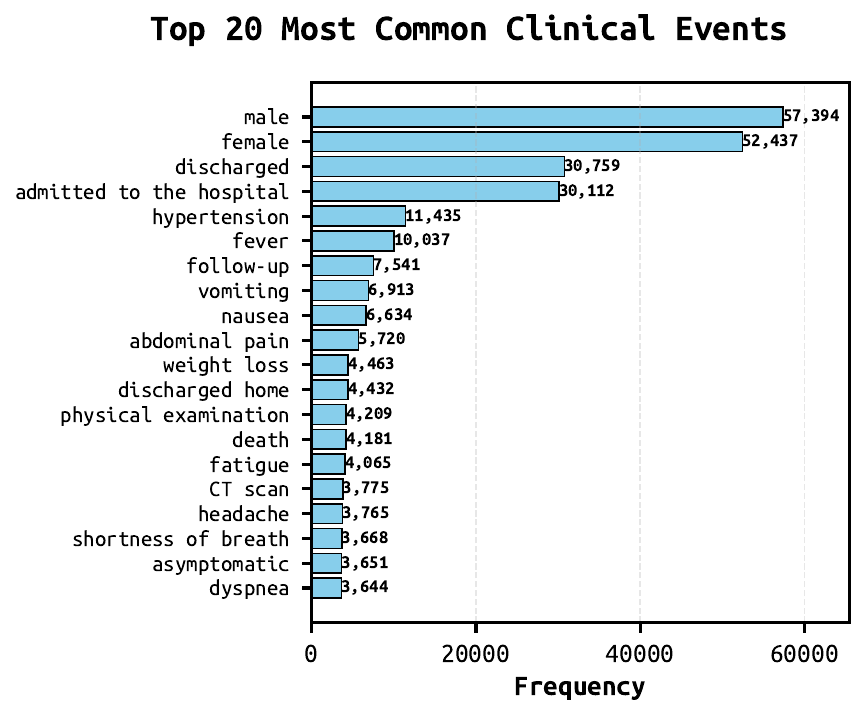}
        \caption*{(b) Most frequently occurring events across all case reports.
        }
    \end{minipage}
    \caption{Overview of textual time series characteristics in PMOA-TTS.}
    \label{fig:tts_stats}
    \vspace{-7.5mm}
\end{figure}

The textual time series (TTS) derived from case reports show substantial variation in sequence length and temporal span. 
On average, each TTS contains $46$ timesteps (mean) %
with 50\% of sequences comprising between 33 and 57 timesteps (Figure~\ref{fig:tts_stats}a). 
The length distribution is right-skewed, with a small number of very long trajectories exceeding $1{,}000$ timesteps. 
Events are generally concise: the mean event span is 3.6 words (26.1 characters), and most events fall between two and five words, such as \texttt{“admitted to the hospital”} or \texttt{“CT scan performed”}. 
Longer spans (e.g., $\geq 15$ words) typically correspond to detailed histopathologic or imaging descriptions.

Figure~\ref{fig:tts_stats}b summarizes the most frequent event strings across the corpus and reflects commonly documented demographic descriptors and clinical milestones (e.g., \texttt{``male''}, \texttt{``female''}, \texttt{``admitted to the hospital''}, \texttt{``discharged''}), as well as prevalent conditions and symptoms (e.g., \texttt{``hypertension''}, \texttt{``fever''}, \texttt{``vomiting''}). 
Importantly, these counts are computed from the \emph{event strings in the TTS} and therefore capture only information that is explicitly stated and extracted as an event in the narrative. 
As a result, demographic events such as \texttt{``male''} and \texttt{``female''} need not sum to the total number of case reports (124k), because some reports ($14\%$) do not explicitly state sex in a way that is extracted as an event.

The temporal coverage of individual case reports also varies widely. 
The mean case duration is 415.9 days (median 104.0 days), with half of cases spanning between 28 and 391 days and the 99th percentile extending to 4{,}380 days (12 years). 
On average, 77.6\% of events share timestamps with at least one other event, indicating frequent temporal co-occurrence within a report. 
The median uniqueness ratio of event timestamps is 0.14, reflecting the common practice of documenting multiple events at the same recorded time (e.g., grouped laboratory results or procedure bundles). 
Approximately 17\% of events have negative timestamps, representing history prior to the chosen reference point (typically admission), such as past diagnoses or prior treatments.
Together, these statistics summarize the density and temporal heterogeneity of clinical event sequences across PMOA-TTS. 

\vspace{-5mm}

\subsection{Diagnosis Distribution Analysis}
\label{diagnosis-distribution-analysis}

To characterize diagnostic coverage, we prompted the LLM to act as an expert physician and list patient-specific diagnoses for each case report. 
The prompt asked the model to extract final diagnostic labels (e.g., ``acne'', ``DRESS syndrome'') rather than isolated clinical findings (e.g., ``rash'', ``leukocytosis''), to place the primary diagnosis first, and to output one diagnosis per line without explanatory text (full prompt in Supplementary Material, Section~\ref{apd:diagnosis-prompt}).

Free-text diagnoses were then standardized to the Unified Medical Language System (UMLS) using the ScispaCy entity linker (\texttt{en\_core\_sci\_lg}) with the UMLS knowledge base. 
For each extracted diagnosis, we selected the highest-ranked UMLS concept under a high-confidence threshold, and retrieved its canonical name for downstream analysis (for further details, refer to Supplementary Material, Section~\ref{apd:umls-mapping}).

\begin{figure}[!htbp]
    \centering
    \begin{subfigure}[t]{0.39\textwidth}
        \centering
        \raisebox{40pt}{\includegraphics[width=0.99\linewidth]{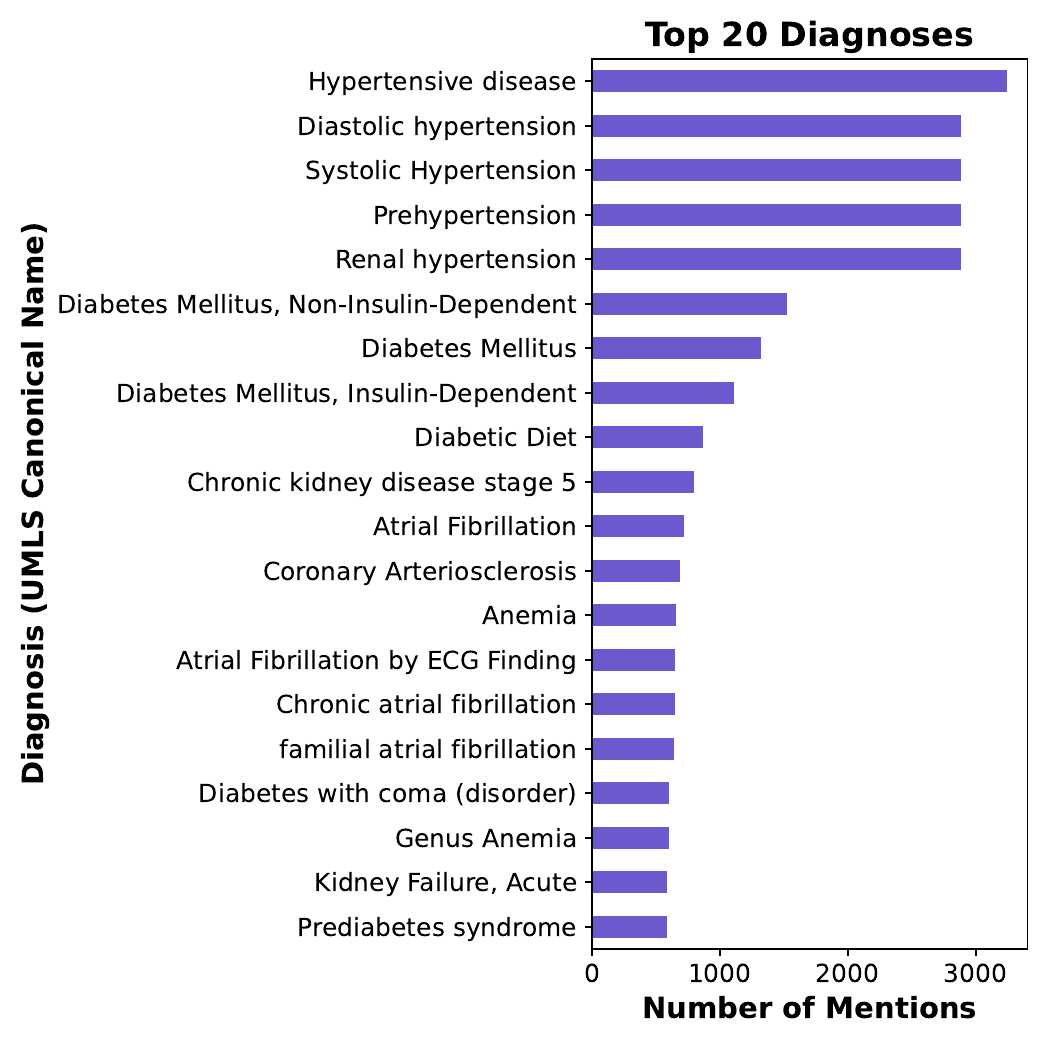}}
        \caption{Most frequent diagnoses.}
        \label{fig:top_diagnoses}
    \end{subfigure}
    \hfill
    \begin{subfigure}[t]{0.6\textwidth}
        \centering
        \includegraphics[width=0.99\linewidth]{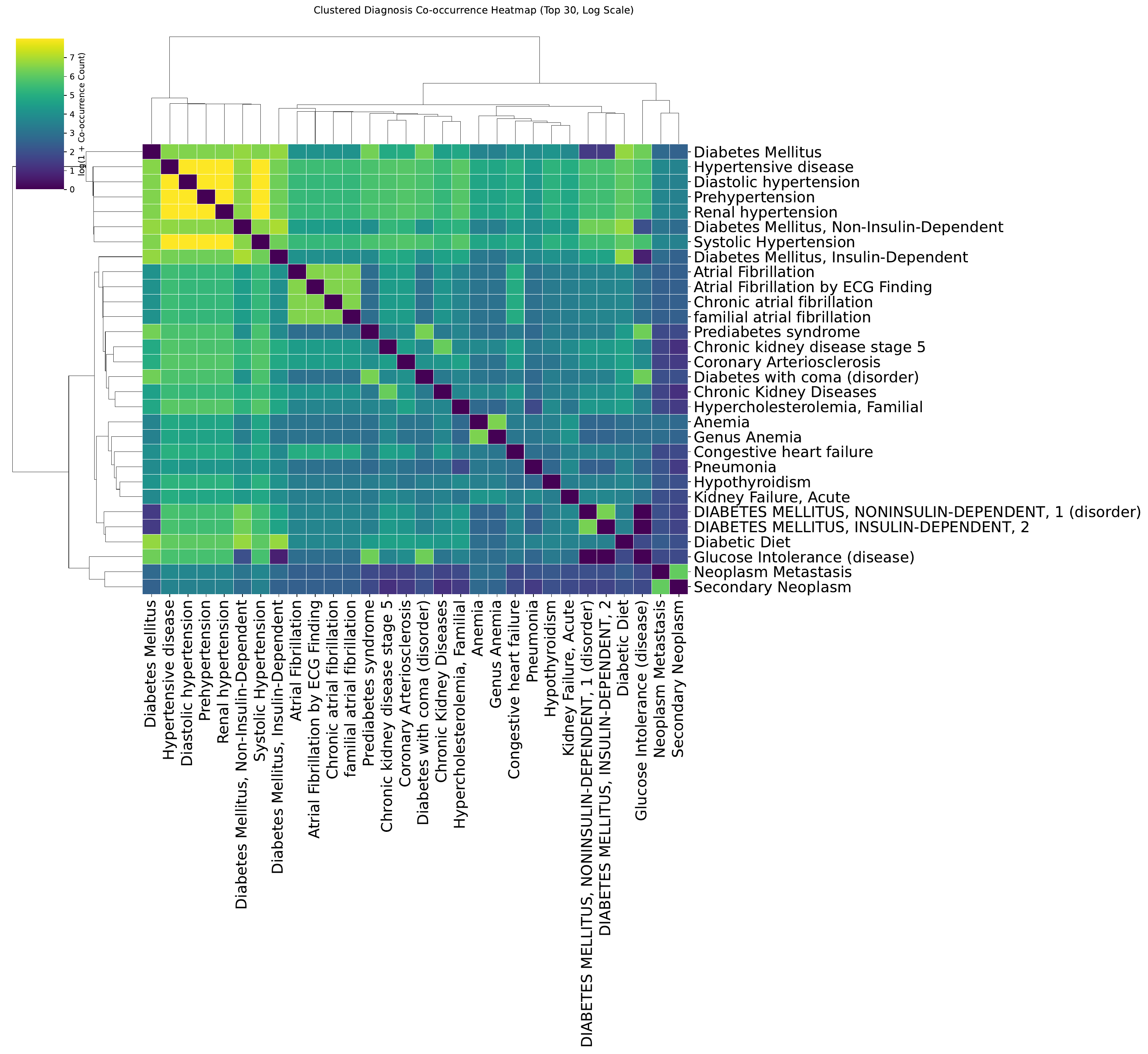}
        \caption{Diagnosis co-occurrence heatmap.}
        \label{fig:diagnosis_cooccurrence}
    \end{subfigure}

    \begin{subfigure}[t]{\textwidth}
        \centering
        \includegraphics[width=0.85\linewidth]{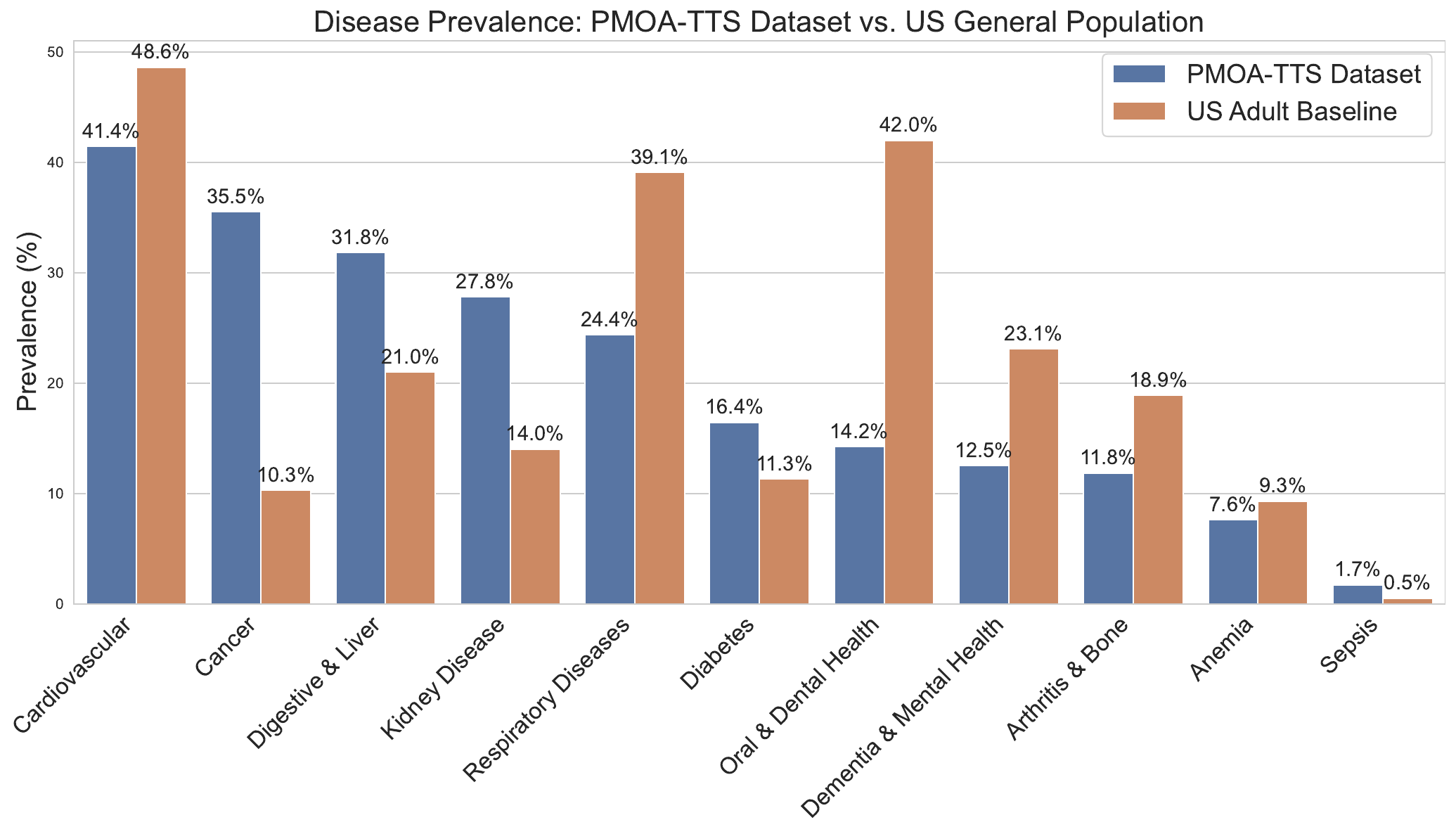}
        \caption{Disease-group prevalence in PMOA-TTS compared with U.S.\ adult baseline.}
        \label{fig:disease_prevalence}
    \end{subfigure}
    \caption{Frequency, co-occurrence, and prevalence patterns of UMLS-normalized diagnoses in PMOA-TTS. (a) The 20 most frequently mentioned diagnoses using canonical UMLS names.
    (b) Pairwise diagnosis co-occurrence (log-transformed counts) with hierarchical clustering, highlighting groups of frequently co-mentioned conditions. (c) Prevalence of coarse disease groups in PMOA-TTS compared with published U.S.\ adult baseline estimates, illustrating systematic differences between case-report-derived narratives
    and general-population distributions.} 
    \label{fig:diagnoses}
    \vspace{-3.5mm}
\end{figure}

Figure~\ref{fig:top_diagnoses} shows the 20 most frequent UMLS-normalized diagnoses in PMOA-TTS. 
These are dominated by common cardiometabolic and renal conditions, including hypertensive disorders (e.g., essential hypertension, systolic and diastolic hypertension), diabetes mellitus and its subtypes, and chronic kidney disease. 
Thus, the most frequently observed diagnoses align with high-prevalence chronic diseases in middle-aged and older adults, consistent with the cohort's age distribution.

We further examined co-occurrence structure using a clustered heatmap of diagnosis pairs (Figure~\ref{fig:diagnosis_cooccurrence}). 
The heatmap reveals coherent clusters of related conditions, such as hypertensive disorders, diabetes variants, and chronic kidney disease, as well as groupings linking cardiovascular and metabolic diagnoses. 
These patterns recapitulate well-established comorbidity relationships (e.g., diabetes with nephropathy, or atrial fibrillation with hypertension) and illustrate how the corpus captures clinically meaningful disease constellations. 

To contextualize diagnosis frequencies beyond listing the most common conditions, we aggregated LLM-extracted, UMLS-normalized diagnoses into coarse disease groups and compared their prevalence in PMOA-TTS with published U.S.\ general-population prevalence estimates (Supplementary Material, Section \ref{apd:umls-mapping}, Table \ref{tab:us_disease_prevalence_sources}). Each case report’s diagnosis set was mapped to a predefined group taxonomy (e.g., cardiovascular, cancer, kidney disease, diabetes) derived from public epidemiologic summaries, and prevalence was computed as the fraction of reports containing at least one diagnosis assigned to each group. These estimates are comparable across groups but are not intended to reflect population-level incidence. To support transparency and reproducibility, we provide the source references used for U.S.\ baseline prevalence estimates and the full mapping procedure in Supplementary Material (Section \ref{apd:umls-mapping}, Table \ref{tab:us_disease_prevalence_sources}).

As shown in Figure~\ref{fig:disease_prevalence}, PMOA-TTS exhibits systematic enrichment of several disease groups relative to U.S.\ baselines, consistent with known publication biases in case reports toward severe, unusual, or diagnostically complex presentations. Cancer-related diagnoses are markedly overrepresented (35.5\% vs.\ 10.3\%), as are kidney disease (27.8\% vs.\ 14.0\%) and digestive/liver conditions (31.8\% vs.\ 21.0\%). In contrast, conditions that are common in routine
care but less frequently the focus of case reports appear underrepresented, including respiratory diseases (24.4\% vs.\ 39.1\%) and oral/dental health (14.2\% vs.\ 42.0\%). Cardiovascular disease remains prevalent in both settings (41.4\% in PMOA-TTS vs.\ 48.6\% baseline), while sepsis,
though rare in the general population, is enriched in PMOA-TTS (1.7\% vs.\ 0.5\%), reflecting the critical-illness emphasis of published case narratives. These comparisons are intended as descriptive characterization rather than epidemiologic inference: PMOA-TTS comprises published case reports rather than a population sample, and prevalence estimates depend on diagnosis-to-group mappings that may be imperfect for some concepts.

\section{Technical Validation}
\label{sec:technical-validation}

\subsection{Case Report Identification Performance}
\label{sec:case-report-identification-performance}

\begin{table}[t!]
\footnotesize
\centering
\caption{Case report identification performance per diagnosis and aggregated across all diagnoses.}
\label{tab:case_report_identification}
\begin{tabular}{llcccc}
\toprule
Subset & Diagnosis & Precision & Recall & Accuracy & F1 Score \\
\midrule
\multirow{6}{*}{PubMed metadata} 
    & Atrial Fibrillation & 0.84 & 0.94 & 0.80 & 0.89 \\
    & COVID-19            & 0.94 & 0.89 & 0.85 & 0.92 \\
    & Diabetes            & 0.85 & 1.00 & 0.85 & 0.92 \\
    & Hypertension        & 1.00 & 1.00 & 1.00 & 1.00 \\
    & Sepsis              & 0.95 & 0.95 & 0.90 & 0.95 \\
\cmidrule(lr){2-6}
    & \textbf{All Diagnoses} & \text{0.92} & \textbf{0.96} & \text{0.88} & \text{0.94} \\
\midrule
\multirow{6}{*}{LLM-Pipeline}
    & Atrial Fibrillation & 1.00 & 1.00 & 1.00 & 1.00 \\
    & COVID-19            & 1.00 & 0.89 & 0.90 & 0.94 \\
    & Diabetes            & 1.00 & 0.94 & 0.95 & 0.97 \\
    & Hypertension        & 1.00 & 0.95 & 0.95 & 0.97 \\
    & Sepsis              & 1.00 & 1.00 & 1.00 & 1.00 \\
\cmidrule(lr){2-6}
    & \textbf{All Diagnoses} & \textbf{1.00} & \textbf{0.96} & \textbf{0.96} & \textbf{0.98} \\
\bottomrule
\end{tabular}
\vspace{-5mm}
\end{table}

Table \ref{tab:case_report_identification} reports the performance of the LLM-based pipeline and a PubMed metadata filter for identifying single-patient case reports on the \texttt{case\_study\_100} benchmark, evaluated across five diagnostic categories using clinician-assigned labels as reference. In this evaluation set, the LLM-based pipeline achieved precision of 1.00 across all diagnostic categories and in aggregate, meaning that every document flagged as a single-patient case report by the pipeline was confirmed as such in this sample. The PubMed metadata filter showed lower and more variable precision, ranging from 0.84 (atrial fibrillation) to 1.00 (hypertension), with an overall precision of 0.92. Both methods achieved similar recall (0.96 overall), indicating comparable ability to recover true case reports.

Overall accuracy was higher for the LLM-based pipeline (0.96 vs.\ 0.88), largely due to better exclusion of multi-patient or non-case-report articles that were sometimes misclassified by the metadata-based filter. Performance patterns were consistent across diagnostic subsets, with the LLM pipeline maintaining high F1 scores (0.94–1.00). These results support the use of the LLM-based approach to construct a high-precision corpus of single-patient narratives for PMOA-TTS, while recognizing that performance is estimated on a finite, diagnosis-targeted benchmark.

\subsection{Evaluation of Clinical Textual Time Series Quality}
\label{sec:evaluation-of-tts-quality}

Using the 200 manually annotated case reports as reference, we evaluated textual time series quality across LLM annotators and runs using event match rate, temporal concordance (c-index), and AULTC 
Table~\ref{tab:model_comparison} reports mean and standard deviation across runs for seven LLM variants: \texttt{Llama 3.3 70B}, \texttt{DeepSeek-R1}, \texttt{DeepSeek-V3-0324}, \texttt{OpenAI GPT5}, \texttt{OpenAI O4-mini}, \texttt{OpenAI O3}, and \texttt{OpenAI O1}. 
\texttt{OpenAI GPT5} achieves the highest mean match rate (0.847 $\pm$ 0.005), indicating the most comprehensive coverage of manually annotated events. \texttt{OpenAI O3} attains the highest concordance (0.891 $\pm$ 0.006), and \texttt{OpenAI O4-mini} yields the highest AULTC (0.748 $\pm$ 0.003), reflecting strong timestamp alignment among matched events. Standard deviations are generally small across runs, suggesting stable behavior for a given model. Similar trends are observed for the LAPJV-based event matching (Supplementary Material, Section \ref{apd:lapjv-matching}, Table \ref{tab:model_comparison_lapjv}).

\begin{table}[h!]
\vspace{-3.5mm}
\footnotesize
\centering
\caption{Comparison of models on Match Rate, Concordance, and AULTC with respect to 200 manually annotated case reports across 5 independent runs for each model.}
\label{tab:model_comparison}
\begin{tabular}{lccc}
\hline
\textbf{Model} & \textbf{Match Rate} & \textbf{Concordance} & \textbf{AULTC} \\
\hline
\texttt{Llama 3.3 70B}       & 0.675 $\pm$ 0.003 & 0.835 $\pm$ 0.003 & 0.717 $\pm$ 0.001 \\
\texttt{DeepSeek-R1}         & 0.746 $\pm$ 0.008 & 0.785 $\pm$ 0.010 & 0.726 $\pm$ 0.003 \\
\texttt{DeepSeek-V3-0324}    & 0.761 $\pm$ 0.004 & 0.791 $\pm$ 0.005 & 0.728 $\pm$ 0.002 \\
\texttt{OpenAI GPT5}         & \textbf{0.847 $\pm$ 0.005} & 0.883 $\pm$ 0.002 & 0.719 $\pm$ 0.001 \\
\texttt{OpenAI O4-mini}      & 0.750 $\pm$ 0.019 & 0.842 $\pm$ 0.008 & \textbf{0.748 $\pm$ 0.003} \\
\texttt{OpenAI O3}           & 0.736 $\pm$ 0.003 & \textbf{0.891 $\pm$ 0.006} & 0.735 $\pm$ 0.003 \\
\texttt{OpenAI O1}           & 0.638 $\pm$ 0.007 & 0.880 $\pm$ 0.003 & 0.730 $\pm$ 0.001 \\
\hline
\end{tabular}
\vspace{-3.5mm}
\end{table}

\begin{table}[ht!]
\vspace{-3.5mm}
\centering
\caption{Bootstrap 95\% confidence intervals for Match Rate, Concordance, and AULTC across models evaluated on the 200 manually annotated case reports.
}
\label{tab:bootstrap_intervals}
\begin{tabular}{lccc}
\hline
\textbf{Model} & \textbf{Match Rate (95\% CI)} & \textbf{Concordance (95\% CI)} & \textbf{AULTC (95\% CI)} \\
\hline
\texttt{Llama 3.3 70B}      & [0.642, 0.718] & [0.833, 0.861] & [0.701, 0.754] \\
\texttt{DeepSeek-R1}        & [0.694, 0.773] & [0.783, 0.818] & [0.708, 0.752] \\
\texttt{DeepSeek-V3-0324}   & [0.727, 0.791] & [0.792, 0.821] & [0.718, 0.758] \\
\texttt{OpenAI GPT5}        & [0.805, 0.866] & [0.882, 0.898] & [0.706, 0.749] \\
\texttt{OpenAI O4-mini}     & [0.692, 0.767] & [0.826, 0.852] & [0.732, 0.769] \\
\texttt{OpenAI O3}          & [0.699, 0.766] & [0.888, 0.907] & [0.728, 0.765] \\
\texttt{OpenAI O1}          & [0.599, 0.677] & [0.872, 0.895] & [0.716, 0.755] \\
\hline
\end{tabular}
\vspace{-3.5mm}
\end{table}

To quantify uncertainty from a complementary perspective, we also performed a bootstrap analysis based on resampling the 200 annotated cases with replacement and recomputing metrics to obtain 95\% confidence intervals. These intervals characterize variability arising from the composition of the reference set, whereas the across-run standard deviations reflect variability in model sampling. The full set of bootstrap intervals is reported in Table~\ref{tab:bootstrap_intervals}.

\begin{figure}[!tbp]
\centering
    \begin{minipage}[t]{0.242\textwidth}
    \centering
    \includegraphics[page=1, width=\linewidth]{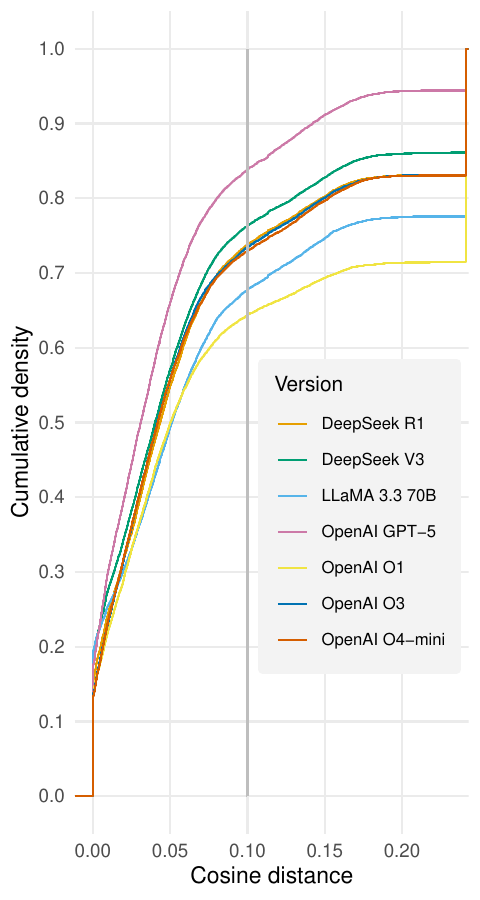}
    \end{minipage}
    \hfill
    \begin{minipage}[t]{0.242\textwidth}
    \centering
    \includegraphics[page=2, width=\linewidth]{Plots_tables/pmoa_updated_man_figs.pdf}
    \end{minipage}
    \hfill
    \begin{minipage}[t]{0.242\textwidth}
    \centering
    \includegraphics[page=3, width=\linewidth]{Plots_tables/pmoa_updated_man_figs.pdf}
    \end{minipage}
    \hfill
    \begin{minipage}[t]{0.242\textwidth}
    \centering
    \includegraphics[page=4, width=\linewidth]{Plots_tables/pmoa_updated_man_figs.pdf}
    \end{minipage}
    \caption{Comparison of different LLM for extracting temporal clinical events from case reports. From left to right: (\textbf{A}) event match cumulative distribution function, (\textbf{B}) concordance box plots, (\textbf{C}) time discrepancy from manual annotation timestamps among matched events (overall), and (\textbf{D}) time discrepancy disaggregated by clinician annotator timestamp (time from presentation). 
    }
    \label{fig:tts_evaluation}
    \vspace{-10pt}
\end{figure}

\subsubsection{Text Event Match Performance}
\label{sec:text-event-match-performance}
Figure~\ref{fig:tts_evaluation} (A) shows the cumulative distributions of cosine distances between model-generated and reference event descriptions, with a threshold of 0.1 used to define a successful semantic match. \texttt{DeepSeek-R1} and its quantized variant \texttt{DeepSeek-R1-UD-IQ1} exhibit the strongest semantic alignment, with the largest fraction
of matched events falling below the threshold. \texttt{DeepSeek-V3-0324} and \texttt{O4-mini} form a second tier, while \texttt{Llama 3.3 70B}, \texttt{O1}, and \texttt{O3} show broader
distributions with greater dispersion. These patterns indicate meaningful differences in how effectively models capture the wording and semantics of clinical event mentions.

\subsubsection{Temporal Ordering Performance}
\label{sec:temporal-ordering-performance}
Using the same similarity threshold to identify matched events, temporal ordering accuracy was evaluated with the concordance index (c-index). As shown in Figure~\ref{fig:tts_evaluation} (B),
multiple models achieve high concordance, indicating that the relative ordering of events is largely preserved once events are successfully aligned. \texttt{Llama 3.3 70B} shows strong
ordering performance despite a lower event match rate, while \texttt{DeepSeek-V3-0324} and \texttt{DeepSeek-R1} combine relatively high concordance with broader event coverage. This
contrast highlights that semantic match quality and temporal ordering accuracy capture distinct aspects of timeline quality: models that align fewer but clearer events may achieve high ordering scores, whereas models with broader coverage face a more challenging ordering task.

\subsubsection{Time Discrepancy Analysis}
\label{sec:time-descrepancy-analysis}
Figure~\ref{fig:tts_evaluation} (C, D) presents cumulative distributions of log-scaled time discrepancies between predicted and reference timestamps. \texttt{DeepSeek-R1} shows the
highest level across discrepancies, indicating closer absolute agreement with manual timestamps, while \texttt{Llama 3.3 70B} exhibits greater dispersion and lower precision. The
interval-stratified analysis in Figure~\ref{fig:tts_evaluation} (D) shows that all models perform well for events near the admission reference time, with accuracy declining as events occur
farther from presentation. Across patient timelines, \texttt{DeepSeek-R1} models yield more precise timestamp estimates, whereas \texttt{Llama 3.3 70B} more consistently preserves relative event ordering. Taken together, these results show that semantic matching, temporal ordering, and
timestamp accuracy provide complementary views of textual time series quality.

\subsection{Additional Sensitivity Analyses}
\label{sec:additional-sensitivity-analyses}

\subsubsection{Varying the Event Match Distance Threshold}

\begin{figure}
    \centering
        \includegraphics[width=\linewidth]{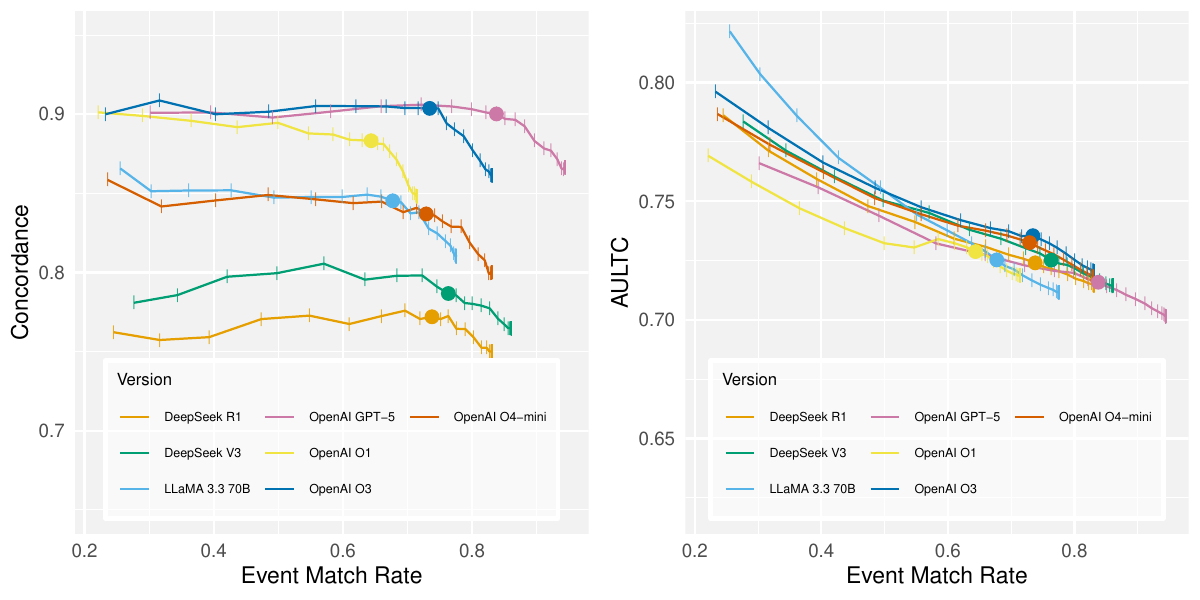}

    \caption{Concordance and AULTC by event match rate for recursive matching heuristic. Solid circle ($\bullet$) represents threshold of 0.1, with ticks (|) indicating 0.01 increments of the threshold in [0.01, 0.50].
    }
    \label{fig:threshold_tradeoff_greedy}
    \vspace{-5mm}
\end{figure}

We conducted sensitivity analyses to assess robustness with respect to the event match threshold. As expected, stricter thresholds produce fewer but cleaner matches, while more permissive thresholds increase coverage at the expense of noisier ordering and timing. Varying the cosine distance threshold from 0.01 to 0.25, we observed corresponding shifts in match rate, concordance, and AULTC. A threshold of 0.1 was chosen based on manual review of \texttt{Llama 3.3 70B Instruct} and \texttt{O1} outputs. The resulting trade-offs are summarized in Figure~\ref{fig:threshold_tradeoff_greedy}.

Similarly, LAPJV preserves high event match rates at low thresholds across models and exhibits a smooth trade-off between coverage and temporal accuracy (Supplementary Material, Section \ref{apd:lapjv-matching} Figure~\ref{fig:threshold_tradeoff_LAPJV}). The resulting operating points are highly comparable to those obtained using the recursive matching heuristic. In particular, the heuristic matcher yields slightly stronger concordance near the elbow of the trade-off curve, while LAPJV achieves marginally higher AULTC values, reflecting improved timestamp alignment under global assignment.  Overall, both strategies lead to consistent qualitative conclusions.

\subsubsection{Sensitivity to Parts of the LLM Prompt Being Changed}
We conducted sensitivity analyses to assess how different components of the extraction prompt affect timeline quality (Supplementary Material, Section~\ref{apd:sensitivity-llm-prompt}). Structured prompts that include explicit temporal instructions (e.g., Interval and Interval+Type) consistently improved semantic alignment, temporal ordering, and timestamp accuracy. In contrast, removing role framing or conjunction expansion led to modest degradation, and Zero-shot prompting performed substantially worse across all metrics. These results highlight that prompt design plays an important role in guiding LLMs to produce high-quality temporal annotations (Figure~\ref{fig:prompt_ablation}).

\subsubsection{Inter-LLM comparison of temporal annotations}
Finally, we qualitatively compared temporal annotations produced by several LLMs on a representative case report to illustrate characteristic differences in event extraction and timestamping (Supplementary Material, Section~\ref{apd:inter-llm-comparison}, Figure~\ref{fig:tts_evaluation}). While all models recovered the major clinical trajectory and preserved temporal polarity, they varied in
event granularity, paraphrasing, and long-range temporal normalization. These differences complement the aggregate quantitative results and highlight the impact of model behavior on the structure of extracted timelines. Common model-specific error patterns, such as granularity variation, duplicate events, and differences in temporal normalization, are summarized in the Supplementary Material, Section~\ref{apd:inter-llm-comparison}, Table~\ref{tab:error_taxonomy}.

\subsection{Limitations and Future Directions}
\label{sec:limitations-and-future-directions}

Our temporal annotation pipeline has several important limitations. First, it operates on case reports from the PubMed Open Access (PMOA) corpus, which often emphasize rare or atypical presentations and differ from routine clinical notes (e.g., progress notes or discharge summaries in MIMIC-III/IV). This may limit direct generalizability to real-world health system documentation. At the same time, case reports offer rich, temporally structured narratives with explicit clinical rationale, making them a useful testbed for developing and stress-testing textual time series methods. Our LLM-based framework is designed to be reusable and, in ongoing work, is being adapted to corpora such as MIMIC-IV discharge summaries for future dataset releases.

Second, we model events as minimally modified text spans aligned to the source narrative. This design choice preserves interpretability and facilitates direct comparison with the underlying text but relies on embedding-based matching for evaluation and does not normalize event content to standard clinical ontologies such as UMLS. Third, representing time with single relative points simplifies annotation and downstream use but cannot fully capture durations or complex temporal relations (e.g., overlapping intervals or recurrent events).

Fourth, our one-to-one recursive event matching strategy emphasizes temporal consistency at the expense of recall, especially when the same clinical concept is expressed multiple times or in conjunctive form.  Alternative matching formulations (e.g., many-to-many or cluster-based matching) could improve coverage but would require more elaborate evaluation protocols. Finally, the manual reference set comprises 200 cases annotated by a single clinical expert, so we cannot estimate inter-annotator agreement, and all metrics reflect agreement with that expert rather than an adjudicated consensus.

These limitations point to several directions for future work, including ontology-aware event normalization, richer temporal representations (intervals and uncertainty), improved matching algorithms, and extension of the pipeline to diverse EHR-derived note types.

\begin{figure}[!thbp]
\centering
    \begin{minipage}[t]{0.242\textwidth}
    \centering
    \includegraphics[page=1, width=\linewidth]{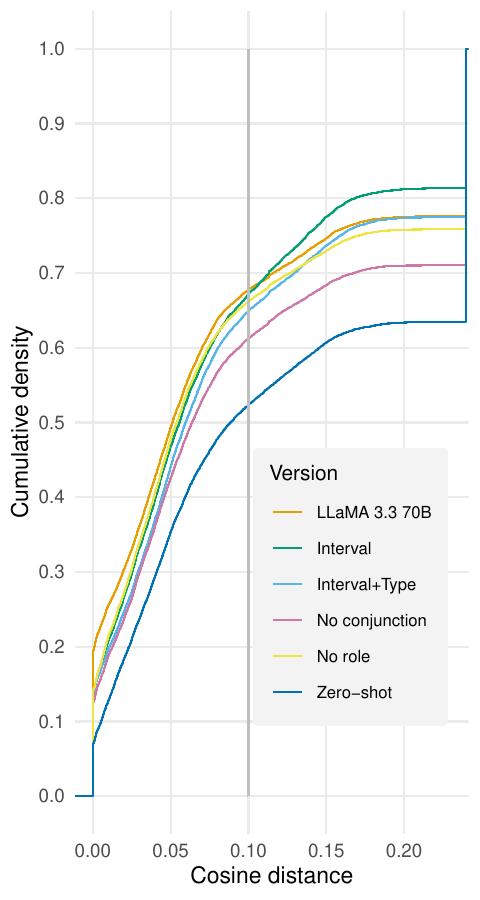}
    \end{minipage}
    \hfill
    \begin{minipage}[t]{0.242\textwidth}
    \centering
    \includegraphics[page=2, width=\linewidth]{Plots_tables/prompt_ablation_updated_figure.pdf}
    \end{minipage}
    \hfill
    \begin{minipage}[t]{0.242\textwidth}
    \centering
    \includegraphics[page=3, width=\linewidth]{Plots_tables/prompt_ablation_updated_figure.pdf}
    \end{minipage}
    \hfill
    \begin{minipage}[t]{0.242\textwidth}
    \centering
    \includegraphics[page=4, width=\linewidth]{Plots_tables/prompt_ablation_updated_figure.pdf}
    \end{minipage}
    \caption{Comparison of \texttt{Llama 3.3 70B} prompt ablation and variant methods for extracting temporal clinical events from case reports. From left to right: (\textbf{A}) event match cumulative distribution function, (\textbf{B}) concordance box plots, (\textbf{C}) time discrepancy from manual annotation timestamps among matched events (overall), and (\textbf{D}) time discrepancy disaggregated by clinician annotator timestamp (time from presentation).}
    \label{fig:prompt_ablation}
    \vspace{-5mm}
\end{figure}

\section{Downstream Application: Survival Analysis}
\label{sec:downstream-application}

We first evaluated whether the extracted textual time series contain information predictive of time to death by formulating a classical survival analysis task. Many case reports specify whether the patient died and, if so, when (e.g., ``the patient died on hospital day 10''), which enables us to define a time-to-event outcome using the \texttt{death\_info} field. For this task, we evaluate models at predefined cut-off times—specifically at 0 hours (admission), 24 hours (1 day), and 168 hours (1 week)—and use the extracted event history up to each cut-off as input. For each case, the sequence of (event, time) pairs is transformed into a fixed-length embedding using a range of pre-trained language models, including encoder-only transformers (e.g., \texttt{bert-base-uncased}, \texttt{roberta-base}, \texttt{ModernBERT}) and larger decoder-style LLMs (e.g., \texttt{Llama 3.3 70B}, \texttt{DeepSeek-R1} and related distilled variants). We additionally include open-source decoder models with documented training corpora (\texttt{OLMo-2-0325-32B-Instruct}, \texttt{RedPajama-INCITE-7B-Instruct}) and a medically fine-tuned decoder model (\texttt{MediPhi-PubMed}) to examine potential domain-specific benefits and mitigate data-leakage concerns. These embeddings are then used as covariates in standard time-to-event models (DeepSurv and DeepHit) trained to predict survival beyond each cut-off. Model performance is summarized using the time-dependent concordance index to assess alignment between predicted and observed survival times. Detailed preprocessing, censoring, and hyperparameter tuning procedures are described in the Supplementary Material, Section~\ref{apd:survival-analysis-pipeline}.

We report results using \texttt{DeepSeek-R1} annotations only (Table~\ref{tab:survival}). Embeddings derived from larger decoder-style LLMs—particularly \texttt{Llama-3.3-70B-Instruct} and \texttt{DeepSeek-R1-Distill-Llama-70B}—yield the highest time-dependent concordance across cut-offs, followed by their smaller variants and then encoder-only baselines such as \texttt{bert-base-uncased}, \texttt{roberta-base}, and \texttt{ModernBERT}. Open-source decoder models with documented training corpora, including \texttt{OLMo-2-0325-32B-Instruct} and \texttt{RedPajama-INCITE-7B-Instruct}, achieve concordance comparable to mid-sized proprietary LLMs, while the medically fine-tuned \texttt{MediPhi-PubMed} performs similarly to or slightly better than general-domain encoders, suggesting a modest benefit from domain-specific adaptation without relying on private EHR data. Across annotation sources and model families, time-dependent concordance values remain well above 0.7 even at the 1-week horizon, indicating that textual representations of clinical event trajectories carry meaningful signal for time-to-death prediction. These results support the viability of text-to-time modeling on PMOA-TTS and highlight that both representation capacity and training data of the underlying language model substantially influence survival analysis performance.

\begin{table*}[t]
\footnotesize
\centering
\begin{adjustbox}{width=\textwidth}
\begin{threeparttable}
\caption{Time-dependent concordance (mean $\pm$ standard deviation across 5 experimental runs) index across embedding models and survival modeling approaches at different time thresholds for DeepSeek-R1 annotations.}
\label{tab:survival}
\begin{tabular}{lcccccc}
\toprule
\multirow{2}{*}{\textbf{Embedding Model}} &
\multicolumn{2}{c}{\textbf{Time Threshold = 0h}} &
\multicolumn{2}{c}{\textbf{Time Threshold = 24h}} &
\multicolumn{2}{c}{\textbf{Time Threshold = 168h}} \\
\cmidrule(lr){2-3} \cmidrule(lr){4-5} \cmidrule(lr){6-7}
 & \textbf{DeepHit} & \textbf{DeepSurv} & 
   \textbf{DeepHit} & \textbf{DeepSurv} &
   \textbf{DeepHit} & \textbf{DeepSurv} \\
\midrule

DeepSeek-R1-Distill-Llama-70B     & 0.815 $\pm$ 0.010 & 0.812 $\pm$ 0.013 & 0.815 $\pm$ 0.012 & 0.818 $\pm$ 0.009 & 0.810 $\pm$ 0.012 & 0.812 $\pm$ 0.014 \\
DeepSeek-R1-Distill-Llama-8B      & 0.806 $\pm$ 0.015 & 0.803 $\pm$ 0.011 & 0.809 $\pm$ 0.013 & 0.806 $\pm$ 0.011 & 0.802 $\pm$ 0.010 & 0.809 $\pm$ 0.011 \\
Llama-3.1-8B-Instruct             & 0.815 $\pm$ 0.008 & 0.813 $\pm$ 0.010 & 0.817 $\pm$ 0.010 & 0.811 $\pm$ 0.007 & 0.808 $\pm$ 0.014 & 0.809 $\pm$ 0.011 \\
Llama-3.3-70B-Instruct            & \textbf{0.822 $\pm$ 0.004} & \textbf{0.817 $\pm$ 0.011} & \textbf{0.821 $\pm$ 0.010} & \textbf{0.819 $\pm$ 0.005} & \textbf{0.819 $\pm$ 0.009} & \textbf{0.822 $\pm$ 0.009} \\
ModernBERT-base                   & 0.718 $\pm$ 0.011 & 0.729 $\pm$ 0.008 & 0.733 $\pm$ 0.010 & 0.738 $\pm$ 0.011 & 0.725 $\pm$ 0.013 & 0.729 $\pm$ 0.009 \\
ModernBERT-large                  & 0.653 $\pm$ 0.014 & 0.667 $\pm$ 0.009 & 0.655 $\pm$ 0.015 & 0.665 $\pm$ 0.017 & 0.643 $\pm$ 0.011 & 0.662 $\pm$ 0.003 \\
roberta-base                      & 0.743 $\pm$ 0.013 & 0.746 $\pm$ 0.008 & 0.742 $\pm$ 0.014 & 0.748 $\pm$ 0.006 & 0.744 $\pm$ 0.009 & 0.747 $\pm$ 0.010 \\
deberta-v3-small                  & 0.739 $\pm$ 0.010 & 0.736 $\pm$ 0.010 & 0.730 $\pm$ 0.019 & 0.758 $\pm$ 0.006 & 0.733 $\pm$ 0.012 & 0.751 $\pm$ 0.006 \\
roberta-base                      & 0.757 $\pm$ 0.017 & 0.756 $\pm$ 0.013 & 0.760 $\pm$ 0.012 & 0.758 $\pm$ 0.014 & 0.761 $\pm$ 0.009 & 0.762 $\pm$ 0.011 \\
BioClinical-ModernBERT-base       & 0.696 $\pm$ 0.013 & 0.708 $\pm$ 0.017 & 0.689 $\pm$ 0.010 & 0.717 $\pm$ 0.013 & 0.713 $\pm$ 0.022 & 0.724 $\pm$ 0.018 \\
BioClinical-ModernBERT-large      & 0.637 $\pm$ 0.012 & 0.646 $\pm$ 0.007 & 0.629 $\pm$ 0.012 & 0.647 $\pm$ 0.023 & 0.626 $\pm$ 0.030 & 0.647 $\pm$ 0.022 \\
MediPhi-PubMed                    & 0.780 $\pm$ 0.015 & 0.738 $\pm$ 0.007 & 0.767 $\pm$ 0.016 & 0.781 $\pm$ 0.009 & 0.782 $\pm$ 0.018 & 0.784 $\pm$ 0.022 \\
OLMo-2-0325-32B-Instruct          & 0.800 $\pm$ 0.011 & 0.797 $\pm$ 0.006 & 0.785 $\pm$ 0.015 & 0.790 $\pm$ 0.021 & 0.789 $\pm$ 0.025 & 0.800 $\pm$ 0.027 \\
RedPajama-INCITE-TB-Instruct      & 0.778 $\pm$ 0.014 & 0.742 $\pm$ 0.015 & 0.771 $\pm$ 0.015 & 0.763 $\pm$ 0.008 & 0.791 $\pm$ 0.019 & 0.798 $\pm$ 0.018 \\
\bottomrule

\end{tabular}
\end{threeparttable}
\end{adjustbox}
\end{table*}

\section{Usage Notes}
\label{sec:usage-notes}

Our full codebase, including preprocessing scripts, prompt templates, and evaluation pipelines,
is available at: \url{https://github.com/jcweiss2/pmoa_tts}. Each subdirectory contains its own README.md with detailed instructions and information. Here is the repository structure
\begin{itemize}
    \item \texttt{make\_tts}: Scripts and notebooks for constructing the PMOA-TTS dataset from case reports.
    \item \texttt{forecasting\_results}: Code for forecasting experiments on the textual time series data.
    \item \texttt{survival\_analyses}: Tools for survival prediction tasks using the extracted timelines.
    \item \texttt{TTS\_evaluation}: Evaluation scripts assessing the quality of the extracted timelines.
\end{itemize}

\noindent Here are step-by-step directions on how to use the dataset

\begin{enumerate}

\item \textbf{Clone the repository:}

{\small
\begin{verbatim}
    git clone https://github.com/jcweiss2/pmoa_tts.git
    cd pmoa_tts
\end{verbatim}
}

\item \textbf{Set up the environment:}

\noindent Ensure you have Python 3.10 or higher installed. It's recommended to use a virtual environment.

{\small
\begin{verbatim}
    conda create -n pmoa_tts python=3.10
    conda activate pmoa_tts
    conda install pandas scikit-learn numpy tensorboard -c conda-forge
    conda install pytorch torchvision torchaudio -c pytorch
    pip install transformers
    pip install argparse
    conda install jupyter -c conda-forge
    pip install sentencepiece
\end{verbatim}
}

\item \textbf{Access the PMOA-TTS dataset.} The dataset is available on Hugging Face. You can download it using the \texttt{datasets} library.

{\small
\begin{verbatim}
    from datasets import load_dataset
    dataset = load_dataset("snoroozi/pmoa-tts")
\end{verbatim}
}

\end{enumerate}

\paragraph{Responsible reuse and cautions}

The release and use of PMOA–TTS raise important considerations. Although LLMs achieve strong performance, they may still generate inaccurate or hallucinated outputs, and any downstream use of these annotations without further validation could introduce bias or error. The extraction of structured diagnoses, demographics, and clinical events from public text—though non-identifiable—could also lead to misinterpretation or reinforce historical biases if not handled carefully. Furthermore, while the dataset is intended for research, there is a risk of misapplication in high-stakes clinical settings without expert oversight. Lastly, the stylistic and structural differences between case reports and real-world EHR notes may limit the generalizability of findings derived from PMOA–TTS. These limitations underscore the importance of using the corpus responsibly, with clear reporting of model limitations and appropriate safeguards when applied in downstream tasks.

\section*{Code availability}\label{sec:data_availability}
In line with the philosophy of reproducible research, all codes used in this paper, including those for data preprocessing and technical validation, are accessible at \url{https://github.com/jcweiss2/pmoa_tts}. The usage instructions and parameter settings for all codes can be found at this link.

\bibliography{references}


\begin{thebibliography}{20}
\ifx \bisbn   \undefined \def \bisbn  #1{ISBN #1}\fi
\ifx \binits  \undefined \def \binits#1{#1}\fi
\ifx \bauthor  \undefined \def \bauthor#1{#1}\fi
\ifx \batitle  \undefined \def \batitle#1{#1}\fi
\ifx \bjtitle  \undefined \def \bjtitle#1{#1}\fi
\ifx \bvolume  \undefined \def \bvolume#1{\textbf{#1}}\fi
\ifx \byear  \undefined \def \byear#1{#1}\fi
\ifx \bissue  \undefined \def \bissue#1{#1}\fi
\ifx \bfpage  \undefined \def \bfpage#1{#1}\fi
\ifx \blpage  \undefined \def \blpage #1{#1}\fi
\ifx \burl  \undefined \def \burl#1{\textsf{#1}}\fi
\ifx \doiurl  \undefined \def \doiurl#1{\url{https://doi.org/#1}}\fi
\ifx \betal  \undefined \def \betal{\textit{et al.}}\fi
\ifx \binstitute  \undefined \def \binstitute#1{#1}\fi
\ifx \binstitutionaled  \undefined \def \binstitutionaled#1{#1}\fi
\ifx \bctitle  \undefined \def \bctitle#1{#1}\fi
\ifx \beditor  \undefined \def \beditor#1{#1}\fi
\ifx \bpublisher  \undefined \def \bpublisher#1{#1}\fi
\ifx \bbtitle  \undefined \def \bbtitle#1{#1}\fi
\ifx \bedition  \undefined \def \bedition#1{#1}\fi
\ifx \bseriesno  \undefined \def \bseriesno#1{#1}\fi
\ifx \blocation  \undefined \def \blocation#1{#1}\fi
\ifx \bsertitle  \undefined \def \bsertitle#1{#1}\fi
\ifx \bsnm \undefined \def \bsnm#1{#1}\fi
\ifx \bsuffix \undefined \def \bsuffix#1{#1}\fi
\ifx \bparticle \undefined \def \bparticle#1{#1}\fi
\ifx \barticle \undefined \def \barticle#1{#1}\fi
\bibcommenthead
\ifx \bconfdate \undefined \def \bconfdate #1{#1}\fi
\ifx \botherref \undefined \def \botherref #1{#1}\fi
\ifx \url \undefined \def \url#1{\textsf{#1}}\fi
\ifx \bchapter \undefined \def \bchapter#1{#1}\fi
\ifx \bbook \undefined \def \bbook#1{#1}\fi
\ifx \bcomment \undefined \def \bcomment#1{#1}\fi
\ifx \oauthor \undefined \def \oauthor#1{#1}\fi
\ifx \citeauthoryear \undefined \def \citeauthoryear#1{#1}\fi
\ifx \endbibitem  \undefined \def \endbibitem {}\fi
\ifx \bconflocation  \undefined \def \bconflocation#1{#1}\fi
\ifx \arxivurl  \undefined \def \arxivurl#1{\textsf{#1}}\fi
\csname PreBibitemsHook\endcsname

\bibitem[\protect\citeauthoryear{Jensen et~al.}{2012}]{jensen2012mining}
\begin{barticle}
\bauthor{\bsnm{Jensen}, \binits{P.B.}},
\bauthor{\bsnm{Jensen}, \binits{L.J.}},
\bauthor{\bsnm{Brunak}, \binits{S.}}:
\batitle{Mining electronic health records: towards better research applications and clinical care}.
\bjtitle{Nature Reviews Genetics}
\bvolume{13}(\bissue{6}),
\bfpage{395}--\blpage{405}
(\byear{2012})
\end{barticle}
\endbibitem

\bibitem[\protect\citeauthoryear{Nor{\'e}n et~al.}{2010}]{noren2010temporal}
\begin{barticle}
\bauthor{\bsnm{Nor{\'e}n}, \binits{G.N.}},
\bauthor{\bsnm{Hopstadius}, \binits{J.}},
\bauthor{\bsnm{Bate}, \binits{A.}},
\bauthor{\bsnm{Star}, \binits{K.}},
\bauthor{\bsnm{Edwards}, \binits{I.R.}}:
\batitle{Temporal pattern discovery in longitudinal electronic patient records}.
\bjtitle{Data Mining and Knowledge Discovery}
\bvolume{20},
\bfpage{361}--\blpage{387}
(\byear{2010})
\end{barticle}
\endbibitem

\bibitem[\protect\citeauthoryear{Johnson et~al.}{2016}]{johnson2016mimic}
\begin{barticle}
\bauthor{\bsnm{Johnson}, \binits{A.E.}},
\bauthor{\bsnm{Pollard}, \binits{T.J.}},
\bauthor{\bsnm{Shen}, \binits{L.}},
\bauthor{\bsnm{Lehman}, \binits{L.-w.H.}},
\bauthor{\bsnm{Feng}, \binits{M.}},
\bauthor{\bsnm{Ghassemi}, \binits{M.}},
\bauthor{\bsnm{Moody}, \binits{B.}},
\bauthor{\bsnm{Szolovits}, \binits{P.}},
\bauthor{\bsnm{Anthony~Celi}, \binits{L.}},
\bauthor{\bsnm{Mark}, \binits{R.G.}}:
\batitle{{MIMIC-III}, a freely accessible critical care database}.
\bjtitle{Scientific Data}
\bvolume{3}(\bissue{1}),
\bfpage{1}--\blpage{9}
(\byear{2016})
\end{barticle}
\endbibitem

\bibitem[\protect\citeauthoryear{Johnson et~al.}{2023}]{johnson2023mimic}
\begin{barticle}
\bauthor{\bsnm{Johnson}, \binits{A.E.}},
\bauthor{\bsnm{Bulgarelli}, \binits{L.}},
\bauthor{\bsnm{Shen}, \binits{L.}},
\bauthor{\bsnm{Gayles}, \binits{A.}},
\bauthor{\bsnm{Shammout}, \binits{A.}},
\bauthor{\bsnm{Horng}, \binits{S.}},
\bauthor{\bsnm{Pollard}, \binits{T.J.}},
\bauthor{\bsnm{Hao}, \binits{S.}},
\bauthor{\bsnm{Moody}, \binits{B.}},
\bauthor{\bsnm{Gow}, \binits{B.}}, \betal:
\batitle{{MIMIC-IV}, a freely accessible electronic health record dataset}.
\bjtitle{Scientific data}
\bvolume{10}(\bissue{1}),
\bfpage{1}
(\byear{2023})
\end{barticle}
\endbibitem

\bibitem[\protect\citeauthoryear{Viani et~al.}{2021}]{viani2021temporal}
\begin{barticle}
\bauthor{\bsnm{Viani}, \binits{N.}},
\bauthor{\bsnm{Kam}, \binits{J.}},
\bauthor{\bsnm{Yin}, \binits{L.}},
\bauthor{\bsnm{Verma}, \binits{S.}},
\bauthor{\bsnm{Stewart}, \binits{R.}},
\bauthor{\bsnm{Patel}, \binits{R.}},
\bauthor{\bsnm{Velupillai}, \binits{S.}}:
\batitle{Temporal information extraction from clinical text}.
\bjtitle{Journal of Biomedical Informatics}
\bvolume{117},
\bfpage{103771}
(\byear{2021})
\end{barticle}
\endbibitem

\bibitem[\protect\citeauthoryear{Cheng and Weiss}{2023}]{cheng2023typed}
\begin{bchapter}
\bauthor{\bsnm{Cheng}, \binits{C.}},
\bauthor{\bsnm{Weiss}, \binits{J.C.}}:
\bctitle{Typed markers and context for clinical temporal relation extraction}.
In: \bbtitle{Machine Learning for Healthcare Conference},
pp. \bfpage{94}--\blpage{109}
(\byear{2023}).
\bcomment{PMLR}
\end{bchapter}
\endbibitem

\bibitem[\protect\citeauthoryear{Sun et~al.}{2013}]{sun2013evaluating}
\begin{barticle}
\bauthor{\bsnm{Sun}, \binits{W.}},
\bauthor{\bsnm{Rumshisky}, \binits{A.}},
\bauthor{\bsnm{Uzuner}, \binits{O.}}:
\batitle{Evaluating temporal relations in clinical text: 2012 i2b2 challenge}.
\bjtitle{Journal of the American Medical Informatics Association}
\bvolume{20}(\bissue{5}),
\bfpage{806}--\blpage{813}
(\byear{2013})
\end{barticle}
\endbibitem

\bibitem[\protect\citeauthoryear{Bethard et~al.}{2016}]{bethard2016semeval}
\begin{bchapter}
\bauthor{\bsnm{Bethard}, \binits{S.}},
\bauthor{\bsnm{Savova}, \binits{G.}},
\bauthor{\bsnm{Chen}, \binits{W.-T.}},
\bauthor{\bsnm{Derczynski}, \binits{L.}},
\bauthor{\bsnm{Pustejovsky}, \binits{J.}},
\bauthor{\bsnm{Verhagen}, \binits{M.}}:
\bctitle{Semeval-2016 task 12: Clinical tempeval}.
In: \bbtitle{Proceedings of the 10th International Workshop on Semantic Evaluation (SemEval-2016)},
pp. \bfpage{1052}--\blpage{1062}
(\byear{2016})
\end{bchapter}
\endbibitem

\bibitem[\protect\citeauthoryear{Galvan et~al.}{2018}]{galvan2018investigating}
\begin{bchapter}
\bauthor{\bsnm{Galvan}, \binits{D.}},
\bauthor{\bsnm{Okazaki}, \binits{N.}},
\bauthor{\bsnm{Matsuda}, \binits{K.}},
\bauthor{\bsnm{Inui}, \binits{K.}}:
\bctitle{Investigating the challenges of temporal relation extraction from clinical text}.
In: \bbtitle{Proceedings of the Ninth International Workshop on Health Text Mining and Information Analysis},
pp. \bfpage{55}--\blpage{64}
(\byear{2018})
\end{bchapter}
\endbibitem

\bibitem[\protect\citeauthoryear{Wang and Weiss}{2025}]{wang2025large}
\begin{bchapter}
\bauthor{\bsnm{Wang}, \binits{J.}},
\bauthor{\bsnm{Weiss}, \binits{J.}}:
\bctitle{A large-language model framework for relative timeline extraction from pubmed case reports}.
In: \bbtitle{Proceedings of the AMIA Informatics Summit}
(\byear{2025}).
\bcomment{American Medical Informatics Association}
\end{bchapter}
\endbibitem

\bibitem[\protect\citeauthoryear{Grattafiori et~al.}{2024}]{grattafiori2024llama}
\begin{botherref}
\oauthor{\bsnm{Grattafiori}, \binits{A.}},
\oauthor{\bsnm{Dubey}, \binits{A.}},
\oauthor{\bsnm{Jauhri}, \binits{A.}},
\oauthor{\bsnm{Pandey}, \binits{A.}},
\oauthor{\bsnm{Kadian}, \binits{A.}},
\oauthor{\bsnm{Al-Dahle}, \binits{A.}},
\oauthor{\bsnm{Letman}, \binits{A.}},
\oauthor{\bsnm{Mathur}, \binits{A.}},
\oauthor{\bsnm{Schelten}, \binits{A.}},
\oauthor{\bsnm{Vaughan}, \binits{A.}}, et al.:
The llama 3 herd of models.
arXiv preprint arXiv:2407.21783
(2024)
\end{botherref}
\endbibitem

\bibitem[\protect\citeauthoryear{Guo et~al.}{2025}]{guo2025deepseek}
\begin{botherref}
\oauthor{\bsnm{Guo}, \binits{D.}},
\oauthor{\bsnm{Yang}, \binits{D.}},
\oauthor{\bsnm{Zhang}, \binits{H.}},
\oauthor{\bsnm{Song}, \binits{J.}},
\oauthor{\bsnm{Zhang}, \binits{R.}},
\oauthor{\bsnm{Xu}, \binits{R.}},
\oauthor{\bsnm{Zhu}, \binits{Q.}},
\oauthor{\bsnm{Ma}, \binits{S.}},
\oauthor{\bsnm{Wang}, \binits{P.}},
\oauthor{\bsnm{Bi}, \binits{X.}}, et al.:
Deepseek-r1: Incentivizing reasoning capability in llms via reinforcement learning.
arXiv preprint arXiv:2501.12948
(2025)
\end{botherref}
\endbibitem

\bibitem[\protect\citeauthoryear{Uzuner et~al.}{2011}]{uzuner20112010}
\begin{barticle}
\bauthor{\bsnm{Uzuner}, \binits{{\"O}.}},
\bauthor{\bsnm{South}, \binits{B.R.}},
\bauthor{\bsnm{Shen}, \binits{S.}},
\bauthor{\bsnm{DuVall}, \binits{S.L.}}:
\batitle{2010 i2b2/{VA} challenge on concepts, assertions, and relations in clinical text}.
\bjtitle{Journal of the American Medical Informatics Association}
\bvolume{18}(\bissue{5}),
\bfpage{552}--\blpage{556}
(\byear{2011})
\end{barticle}
\endbibitem

\bibitem[\protect\citeauthoryear{Jonker and Volgenant}{1987}]{jonker1987shortest}
\begin{barticle}
\bauthor{\bsnm{Jonker}, \binits{R.}},
\bauthor{\bsnm{Volgenant}, \binits{A.}}:
\batitle{A shortest augmenting path algorithm for dense and sparse linear assignment problems}.
\bjtitle{Computing}
\bvolume{38}(\bissue{4}),
\bfpage{325}--\blpage{340}
(\byear{1987})
\end{barticle}
\endbibitem

\bibitem[\protect\citeauthoryear{Tsao et~al.}{2023}]{tsao2023heart}
\begin{barticle}
\bauthor{\bsnm{Tsao}, \binits{C.W.}},
\bauthor{\bsnm{Aday}, \binits{A.W.}},
\bauthor{\bsnm{Almarzooq}, \binits{Z.I.}},
\bauthor{\bsnm{Anderson}, \binits{C.A.}},
\bauthor{\bsnm{Arora}, \binits{P.}},
\bauthor{\bsnm{Avery}, \binits{C.L.}},
\bauthor{\bsnm{Baker-Smith}, \binits{C.M.}},
\bauthor{\bsnm{Beaton}, \binits{A.Z.}},
\bauthor{\bsnm{Boehme}, \binits{A.K.}},
\bauthor{\bsnm{Buxton}, \binits{A.E.}}, \betal:
\batitle{Heart disease and stroke statistics—2023 update: a report from the {American} {Heart} {Association}}.
\bjtitle{Circulation}
\bvolume{147}(\bissue{8}),
\bfpage{93}--\blpage{621}
(\byear{2023})
\end{barticle}
\endbibitem

\bibitem[\protect\citeauthoryear{Pleasants et~al.}{2022}]{pleasants2022respiratory}
\begin{barticle}
\bauthor{\bsnm{Pleasants}, \binits{R.A.}},
\bauthor{\bsnm{Heidari}, \binits{K.}},
\bauthor{\bsnm{Ohar}, \binits{J.}},
\bauthor{\bsnm{Donohue}, \binits{J.F.}},
\bauthor{\bsnm{Lugogo}, \binits{N.L.}},
\bauthor{\bsnm{Kanotra}, \binits{S.M.}},
\bauthor{\bsnm{Kraft}, \binits{M.}},
\bauthor{\bsnm{Mannino}, \binits{D.M.}},
\bauthor{\bsnm{Strange}, \binits{C.B.}}:
\batitle{Respiratory symptoms among us adults: a cross-sectional health survey study}.
\bjtitle{Pulmonary Therapy}
\bvolume{8}(\bissue{3}),
\bfpage{255}--\blpage{268}
(\byear{2022})
\end{barticle}
\endbibitem

\bibitem[\protect\citeauthoryear{Abu-Tineh et~al.}{2023}]{abu2023rare}
\begin{barticle}
\bauthor{\bsnm{Abu-Tineh}, \binits{M.}},
\bauthor{\bsnm{Alamin}, \binits{M.A.}},
\bauthor{\bsnm{Aljaloudi}, \binits{E.}},
\bauthor{\bsnm{Alshurafa}, \binits{A.}},
\bauthor{\bsnm{Garcia-Ca{\~n}ibano}, \binits{B.}},
\bauthor{\bsnm{Taha}, \binits{R.Y.}},
\bauthor{\bsnm{Elkourashy}, \binits{S.A.}}:
\batitle{A rare case of lambert-eaton myasthenia syndrome associated with non-hodgkin’s lymphoma: A case report and review of the literature}.
\bjtitle{Case Reports in Oncology}
\bvolume{16}(\bissue{1}),
\bfpage{1300}--\blpage{1305}
(\byear{2023})
\end{barticle}
\endbibitem

\bibitem[\protect\citeauthoryear{Katzman et~al.}{2018}]{katzman2018deepsurv}
\begin{barticle}
\bauthor{\bsnm{Katzman}, \binits{J.L.}},
\bauthor{\bsnm{Shaham}, \binits{U.}},
\bauthor{\bsnm{Cloninger}, \binits{A.}},
\bauthor{\bsnm{Bates}, \binits{J.}},
\bauthor{\bsnm{Jiang}, \binits{T.}},
\bauthor{\bsnm{Kluger}, \binits{Y.}}:
\batitle{{DeepSurv}: personalized treatment recommender system using a {Cox} proportional hazards deep neural network}.
\bjtitle{BMC Medical Research Methodology}
\bvolume{18},
\bfpage{1}--\blpage{12}
(\byear{2018})
\end{barticle}
\endbibitem

\bibitem[\protect\citeauthoryear{Lee et~al.}{2018}]{lee2018deephit}
\begin{bchapter}
\bauthor{\bsnm{Lee}, \binits{C.}},
\bauthor{\bsnm{Zame}, \binits{W.}},
\bauthor{\bsnm{Yoon}, \binits{J.}},
\bauthor{\bsnm{Van Der~Schaar}, \binits{M.}}:
\bctitle{{DeepHit}: A deep learning approach to survival analysis with competing risks}.
In: \bbtitle{Proceedings of the AAAI Conference on Artificial Intelligence}
(\byear{2018})
\end{bchapter}
\endbibitem

\bibitem[\protect\citeauthoryear{Antolini et~al.}{2005}]{antolini2005time}
\begin{barticle}
\bauthor{\bsnm{Antolini}, \binits{L.}},
\bauthor{\bsnm{Boracchi}, \binits{P.}},
\bauthor{\bsnm{Biganzoli}, \binits{E.}}:
\batitle{A time-dependent discrimination index for survival data}.
\bjtitle{Statistics in Medicine}
\bvolume{24}(\bissue{24}),
\bfpage{3927}--\blpage{3944}
(\byear{2005})
\end{barticle}
\endbibitem

\end{thebibliography}

\clearpage
\appendix
\begin{center}
    {\LARGE Supplementary Material: Temporally annotated textual time series from PubMed Open Access clinical case reports}
\end{center}
\vspace{1em}

\section{LLM Prompts}
\label{apd:llm-prompts}

\subsection{LLM Prompt to Generate Textual Time Series from PMOA Case Reports}
\label{apd:extraction-prompt}

\begin{tcolorbox}[
  colback=gray!5,
  colframe=gray!100,
  title=Prompt,
  boxrule=2pt,
  enhanced,
  breakable,                %
  before skip=8pt,          %
  after skip=8pt,           %
  pad at break*=2mm,        %
  fontupper=\scriptsize,
  title after break={Prompt} %
]
You are a physician. Extract the clinical events and the related time stamp from the case report. The admission event has timestamp 0. If the event is not available, we treat the event, e.g. current main clinical diagnosis or treatment with timestamp 0. The events that happened before event with 0 timestamp have negative time, the ones after the event with 0 timestamp have positive time. The timestamp are in hours. The unit will be omitted when output the result. If there is no temporal information of the event, please use your knowledge and events with temporal expression before and after the events to provide an approximation. We want to predict the future events given the events happened in history. For example, here is the case report.\\

\textcolor{teal}{\texttt{An 18-year-old male was admitted to the hospital with a 3-day history of fever and rash. Four weeks ago, he was diagnosed with acne and received the treatment with minocycline, 100 mg daily, for 3 weeks. With increased WBC count, eosinophilia, and systemic involvement, this patient was diagnosed with DRESS syndrome. The fever and rash persisted through admission, and diffuse erythematous or maculopapular eruption with pruritus was present. One day later the patient was discharged.}}\\

Let's find the locations of events in the case report, it shows that four weeks ago of fever and rash, four weeks ago, he was diagnosed with acne and receive treatment. So the event of fever and rash happened four weeks ago, 672 hours, it is before admitted to the hospital, so the time stamp is -672. diffuse erythematous or maculopapular eruption with pruritus was documented on the admission exam, so the timestamp is 0 hours, since it happens right at admission. DRESS syndrome has no specific time, but it should happen soon after admission to the hospital, so we use our clinical judgment to give the diagnosis of DRESS syndrome the timestamp 0. then the output should look like:\\

\texttt{18 years old | 0 \\
male | 0 \\
admitted to the hospital | 0
fever | -72 \\
rash | -72 \\
acne | -672 \\
minocycline | -672 \\
increased WBC count | 0 \\
eosinophilia| 0 \\
systemic involvement| 0 \\
diffuse erythematous or maculopapular eruption| 0 \\
pruritis | 0 \\
DRESS syndrome | 0 \\
fever persisted | 0 \\
rash persisted | 0 \\
discharged | 24} \\

Separate conjunctive phrases into their component events and assign them the same timestamp (for example, separation of ‘fever and rash’ into 2 events: ‘fever’ and ‘rash’). If the event has duration, assign the event time as the start of the time interval. Attempt to use the text span without modifications except ‘history of’ where applicable. Include all patient events, even if they appear in the discussion; do not omit any events; include termination/discontinuation events; include the pertinent negative findings, like ‘no shortness of breath’ and ‘denies chest pain’. Show the events and timestamps in rows, each row has two columns: one column for the event, the other column for the timestamp. The time is a numeric value in hour unit. The two columns are separated by a pipe ‘|’ as a bar-separated file. Reply with the table only.
\end{tcolorbox}

\subsection{LLM Prompt to Extract Demographics from PMOA Case Reports}
\label{apd:demographics-prompt}
\begin{tcolorbox}[
  colback=gray!5,
  colframe=gray!100,
  title={Prompt},
  boxrule=2pt,
  enhanced,
  breakable,                %
  before skip=8pt,          %
  after skip=8pt,           %
  pad at break*=2mm,        %
  fontupper=\scriptsize,
  title after break={{Prompt}} %
]
You are an expert clinician. Extract the following demographic information from the case report: (1) age at case presentation, (2) sex (biologic sex at birth), and (3) ethnicity.  Report the age in years.  Report the sex as Male, Female, otherwise as a custom string, or Not Specified.  Report ethnicity as according to the US Census, otherwise as Not Specified.  For example, here is the case report.\\

\textcolor{teal}{\texttt{An 18-year-old male was admitted to the hospital with a 3-day history of fever and rash. Four weeks ago, he was diagnosed with acne and received the treatment with minocycline, 100 mg daily, for 3 weeks. With increased WBC count, eosinophilia, and systemic involvement, this patient was diagnosed with DRESS syndrome. The fever and rash persisted through admission, and diffuse erythematous or maculopapular eruption with pruritus was present. One day later the patient was discharged.}}\\

Then the demographic extraction should be:\\
\texttt{18 | Male | Not Specified} \\
 
Report the demographics with the three columns separated by a pipe '|' as a bar-separated row as above.
\end{tcolorbox}

\subsection{LLM Prompt to Extract List of Diagnoses from PMOA Case Reports}
\label{apd:diagnosis-prompt}
\begin{tcolorbox}[
  colback=gray!5,
  colframe=gray!100,
  title=LLM prompt to extract list of diagnoses from PMOA case reports,
  boxrule=2pt,
  enhanced,
  breakable,                %
  before skip=8pt,          %
  after skip=8pt,           %
  pad at break*=2mm,        %
  fontupper=\scriptsize,
  title after break={Prompt} %
]
\textcolor{teal}{\texttt{An 18-year-old male was admitted to the hospital with a 3-day history of fever and rash. Four weeks ago, he was diagnosed with acne and received the treatment with minocycline, 100 mg daily, for 3 weeks. With increased WBC count, eosinophilia, and systemic involvement, this patient was diagnosed with DRESS syndrome. The fever and rash persisted through admission and diffuse erythematous or maculopapular eruption with pruritus was present. One day later the patient was discharged.}} \\

Then the output should look like:\\
    \texttt{acne} \\
    \texttt{DRESS syndrome} \\

Rash, leukocytosis, and other findings are not diseases so they are omitted, while acne and DRESS syndrome were both diagnosed, so they are included. Output a list of diagnoses (one per line) from the following case. Place the primary diagnosis first. Include only the list and nothing else.

\end{tcolorbox}

\section{Evaluation Metrics}
\label{apd:evaluation-metrics}

We evaluated textual time series extracted from PMOA case reports along three dimensions:
(i) semantic alignment between predicted and manually annotated events (event match rate),
(ii) agreement in temporal ordering (temporal concordance), and
(iii) agreement in timestamp values (time discrepancy).  
Each metric captures a distinct property of timeline quality.

\subsection{Event Match Rate}
\label{apd:event-match-rate}

To assess how well predicted clinical events align with reference events, we use a recursive
best-match strategy adapted from prior work~\cite{wang2025large} as shown in Algorithm~\ref{alg:recursive_match}. At each step, the method
identifies the closest unmatched pair of predicted and reference events using a text-similarity
metric, retains the pair if it satisfies a distance threshold, and removes both events before
proceeding. \textbf{Algorithm \ref{alg:recursive_match}} provides pseudocode for this recursive matching procedure.
This strategy produces a one-to-one alignment between reference and predicted events that is
efficient and well-suited to timelines of differing lengths. 

In our final evaluation, we also implement a globally optimal one-to-one matching procedure based on the Linear Assignment Problem, solved using the Jonker–Volgenant (LAPJV) algorithm \citep{jonker1987shortest}. We construct a rectangular cost matrix of pairwise cosine distances between predicted and reference events and enforce a distance threshold by assigning infeasible pairs a prohibitively large cost. LAPJV then computes a minimum-cost assignment under these constraints, yielding a globally optimal one-to-one alignment. This provides a principled benchmark against the recursive heuristic while preserving the same matching constraints. Unlike the recursive best-match heuristic, which greedily selects local minima, the LAPJV formulation optimizes the full matching jointly, ensuring global optimality under the same one-to-one and distance constraints. A detailed sensitivity analysis of LAPJV-based matching with respect to the distance threshold,
including trade-offs between event match rate, temporal concordance, and timestamp accuracy, is presented in Appendix~\ref{apd:lapjv-matching}.

We evaluated several similarity measures—including Levenshtein distance, BERT-based
embeddings, and PubMedBERT embeddings—and found cosine similarity using PubMedBERT
sentence embeddings to perform best. A cosine distance threshold of 0.1 is used to determine
whether a predicted event is considered a semantic match.

Under this procedure, the event match rate is defined as:
\[
\text{Match Rate} =
\frac{\#\{\text{reference events with a matched prediction}\}}
     {\#\{\text{reference events}\}},
\]
which measures the proportion of reference events that are successfully recovered.

\begin{algorithm}[ht]
\caption{Recursive Best Match}
\scriptsize
\label{alg:recursive_match}
\begin{algorithmic}[1]
\Require Two lists: \texttt{ref} (manually annotated reference events) and \texttt{pred} (predicted events)
\Ensure List of best-matching event pairs
\Function{MatchEvents}{\texttt{ref}, \texttt{pred}}
    \If{\texttt{ref} is empty \textbf{or} \texttt{pred} is empty}
        \State \Return \texttt{[]}
    \EndIf
    \State $\texttt{min\_distance} \gets \infty$
    \State $\texttt{best\_pair} \gets \texttt{None}$
    \ForAll{$r \in \texttt{ref}$}
        \ForAll{$p \in \texttt{pred}$}
            \State $d \gets \texttt{ComputeDistance}(r, p)$
            \If{$d < \texttt{min\_distance}$}
                \State $\texttt{min\_distance} \gets d$
                \State $\texttt{best\_pair} \gets (r, p)$
            \ElsIf{$d = \texttt{min\_distance}$}
                \State $(r^*, p^*) \gets \texttt{best\_pair}$
                \State Get indices (within the annotation files) of $r$, $p$, and also of $r^*, p^*$
                \If{index of $r <$ index of $r^*$}
                    \State $\texttt{best\_pair} \gets (r, p)$
                \ElsIf{index of $r = $ index of $r^*$ \textbf{and} index of $p <$ index of $p^*$}
                    \State $\texttt{best\_pair} \gets (r, p)$
                \EndIf
            \EndIf
        \EndFor
    \EndFor
    \State $(r^*,p^*) \gets \texttt{best\_pair}$
    \State Remove $r^*$ from \texttt{ref}
    \State Remove $p^*$ from \texttt{pred}
    \State $\texttt{result} \gets \texttt{[best\_pair]} + \texttt{MatchEvents}(\texttt{ref}, \texttt{pred})$
    \State \Return $\texttt{result}$
\EndFunction
\end{algorithmic}
\end{algorithm}

\subsection{Temporal Concordance}
\label{apd:temporal-concordance}

Temporal ordering accuracy is quantified using the concordance index (c-index), which
measures the probability that pairs of matched events are ordered correctly in predicted time.
Let $t^{\text{ref}}_i$ and $t^{\text{pred}}_i$ denote the reference and predicted timestamps for
matched event $i$. The c-index is:
\[
\text{c-index} = \frac{1}{N}
\sum_{\substack{i<j\\
t^{\text{ref}}_i \neq t^{\text{ref}}_j\\
t^{\text{pred}}_i \neq t^{\text{pred}}_j}}
\mathds{1}\!\left\{
(t^{\text{ref}}_i - t^{\text{ref}}_j)(t^{\text{pred}}_i - t^{\text{pred}}_j) > 0
\right\},
\]
where $N$ is the number of comparable pairs.  
Higher values indicate stronger preservation of reference ordering.

\subsection{Time Discrepancy and AULTC}
\label{apd:aultc}

For each matched event, we measure timestamp accuracy using the absolute time error
$\Delta t_i = |t^{\text{pred}}_i - t^{\text{ref}}_i|$.  
Because timestamp discrepancies can vary across several orders of magnitude, we evaluate
errors on a log scale:
\[
x_i = \log(1 + \Delta t_i).
\]

To characterize timestamp accuracy across the dataset, we compute the empirical CDF (constructed over log-time discrepancies pooled across all matched events in the 200-case gold standard):
\[
F(x) = \frac{1}{k} \sum_{i=1}^{k} \mathds{1}\{x_i \le x\},
\]
where $k$ is the total number of matched events across all annotated cases.

We then summarize the overall discrepancy using the Area Under the Log-Time CDF (AULTC):
\[
\text{AULTC} =
\frac{1}{\log(1+S_{\max})}
\left[
\sum_{i=1}^{k}(x_{(i)} - x_{(i-1)})\frac{i}{k}
+
\big( \log(1+S_{\max}) - x_{(k)} \big)
\right],
\]
where $x_{(i)}$ are the sorted log discrepancies and $S_{\max}$ is the maximum observed
absolute error.  
AULTC ranges from 0 to 1, with higher values indicating closer alignment between predicted
and reference timestamps.

Finally, we stratify timestamp errors by temporal distance (e.g., within 1 hour, 1 day, 1 week,
1 year) to examine how accuracy varies across clinically meaningful time scales.

\section{UMLS Mapping for Diagnosis Standardization}
\label{apd:umls-mapping}

To standardize the diagnoses extracted by LLMs from the case reports, we mapped the free-text diagnoses to concepts in the Unified Medical Language System (UMLS). 
This process involved the following steps:

\begin{enumerate}
    \item \textbf{Initial Extraction}: For each case report, we prompted the LLM to extract diagnoses. 
    This yielded a list of free-text diagnoses for each case report.
    \item \textbf{Entity Linking}: We used ScispaCy's entity linker model 
    \texttt{(en\_core\_sci\_lg)} with the UMLS knowledge base to map each extracted diagnosis to corresponding UMLS Concept Unique Identifiers (CUIs). 
    The entity linker identifies potential UMLS concepts and ranks them based on contextual relevance.
    \item \textbf{Filtering and Disambiguation}: For each diagnosis, we:
    \begin{itemize}
       \item Retained only mappings with a confidence score above 0.85
       \item When multiple CUIs were identified for a single diagnosis, selected the highest-ranked mapping
    \end{itemize}
    \item \textbf{Concept Normalization}: After mapping to CUIs, we:
\begin{itemize}
   \item Retrieved the canonical names from UMLS Metathesaurus
\end{itemize}
\end{enumerate}

\vspace{-3.5mm}
\begin{table}[!htbp]
\footnotesize
\centering
\caption{U.S.\ baseline prevalence estimates for disease groups used in the PMOA--TTS prevalence comparison.}
\label{tab:us_disease_prevalence_sources}
\begin{tabular}{p{3.6cm} p{2.2cm} p{7.8cm}}
\toprule
\textbf{Disease Group} & \textbf{U.S.\ Prevalence} & \textbf{Source and Description} \\
\midrule

\textbf{Cardiovascular} & \textbf{48.6\%} &
American Heart Association Heart Disease and Stroke Statistics~\cite{tsao2023heart}.
Includes hypertension, coronary heart disease, heart failure, and stroke. \\[5pt]

\textbf{Oral \& Dental Health} & \textbf{42.0\%} &
CDC periodontal disease estimates for U.S.\ adults aged $\geq$30 \url{https://www.cdc.gov/oral-health/about/gum-periodontal-disease.html} \\[5pt]

\textbf{Respiratory Diseases} & \textbf{39.1\%} &
Prevalence of respiratory impairment in U.S.\ adults~\cite{pleasants2022respiratory}. \\[5pt]

\textbf{Dementia \& Mental Health} & \textbf{23.1\%} &
Prevalence of any mental illness among U.S.\ adults. \url{https://www.nimh.nih.gov/health/statistics/mental-illness} \\[5pt]

\textbf{Digestive \& Liver} & \textbf{21.0\%} &
Estimated prevalence of digestive diseases in the U.S. \url{https://www.niddk.nih.gov/health-information/health-statistics/digestive-diseases} \\[5pt]

\textbf{Arthritis \& Bone} & \textbf{18.9\%} &
Diagnosed arthritis among U.S.\ adults aged $\geq$18 \url{https://www.cdc.gov/nchs/products/databriefs/db497.htm}. \\[5pt]

\textbf{Kidney Disease} & \textbf{14.0\%} &
Estimated prevalence of chronic kidney disease stages 1--4 \url{https://www.cdc.gov/kidney-disease/php/data-research/index.html}. \\[5pt]

\textbf{Diabetes} & \textbf{11.3\%} &
Diagnosed diabetes prevalence in U.S.\ adults \url{https://www.cdc.gov/nchs/products/databriefs/db516.htm}. \\[5pt]

\textbf{Anemia} & \textbf{9.3\%} &
Prevalence of anemia in the U.S.\ population aged $\geq$2 \url{https://www.cdc.gov/nchs/products/databriefs/db519.htm}. \\[5pt]

\textbf{Cancer} & \textbf{10.3\%} &
Prevalence of cancer survivors in the U.S.\ population \url{https://www.cdc.gov/nchs/fastats/cancer.htm}. \\[5pt]

\textbf{Sepsis} & \textbf{0.5\%} &
Estimated from annual U.S.\ sepsis incidence relative to population size \url{https://www.cdc.gov/sepsis/about/index.html}. \\

\bottomrule
\end{tabular}
\vspace{-3.5mm}
\end{table}

This mapping process allowed us to standardize the diverse expressions of diagnoses extracted by LLMs into a consistent terminology system, enabling more reliable frequency analysis and co-occurrence patterns. 
Across the full PMOA-TTS corpus, LLM-extracted diagnoses yielded
$N_{\text{raw}} = 63{,}715$ unique free-text diagnosis strings.
Of these, $N_{\text{mapped}} = 41{,}216$ ($64.7\%$) were successfully mapped to at least one
UMLS CUI with a confidence score $\ge 0.85$.
When weighted by frequency of occurrence, mappings at this threshold covered
$91.7\%$ of all diagnosis mentions, indicating that commonly occurring clinical
diagnoses were more likely to receive high-confidence mappings.
Mapping coverage decreased monotonically with increasing confidence threshold.
We therefore selected a threshold of 0.85 as a conservative operating point for
downstream normalization, providing broad coverage of frequently occurring diagnoses
while maintaining a high-confidence mapping criterion.

\section{Sensitivity Analyses}
\label{apd:sensitivity-analyses}

\subsection{Sensitivity to Parts of the LLM Prompt Being Changed}
\label{apd:sensitivity-llm-prompt}

We slightly modified the prompt from the preceding section to check for sensitivity to specific parts of the LLM prompt being removed or modified.
This led to the following variants:
\begin{enumerate}
    \item \textbf{No changes}: Original LLM query prompt. %
    
    \item \textbf{No role:} Remove the initial sentence \texttt{"You are a physician."}
    
    \item \textbf{Zero-shot prompting:} Remove the example case report and textual time series from the prompt. \\
    \textcolor{teal}{\texttt{An 18-year-old male was admitted to ...}}\\
    "Let's find the locations of events in the case report, it shows that ... then the output should look like:"\\
    \texttt{18 years old | 0 \\
    male | 0 \\
    ..}

    \item \textbf{No conjunction expansion:} Remove the following sentence from the prompt: \texttt{"Separate conjunctive phrases into their component events and assign them the same timestamp (for example, separation of ‘fever and rash’ into 2 events: ‘fever’ and ‘rash’)"}
    
    \item \textbf{Interval:} A query was added to the prompt for extracting the time interval (start and end time of clinical events). The original prompt’s instructions and few-shot examples were modified accordingly. 
    
    \item \textbf{Interval + Type:} A query was added to the prompt for extracting the time interval and i2b2
    event type (Interval+Type), where event type is one of: Factual, Possible, Hypothetical, Conditional, Negated, Historical, Uncertain. The original prompt’s instructions and few-shot examples were modified accordingly. 
\end{enumerate}

\paragraph{Semantic similarity:}
The cumulative distributions of cosine distances between predicted and reference event phrases show clear differences across prompt variants. \emph{Interval} and \emph{Interval+Type} produce the closest matches, while \emph{Zero-shot} performs substantially worse, with a right-shifted distribution indicating lower-quality event spans. \emph{No role} and \emph{No conjunction} show modest degradation relative to the full prompt. These results
demonstrate that including structured temporal guidance and explicit event formatting improves semantic alignment.

\paragraph{Temporal ordering.}
The concordance index (c-index) follows a similar pattern. \emph{Interval+Type} achieves the highest median concordance, followed by \emph{Interval} and the full prompt. Removing role
framing or conjunction splitting reduces ordering consistency, and \emph{Zero-shot} again yields the lowest performance with greater variability. This suggests that structured instructions help the model maintain coherent temporal ordering among extracted events.

\paragraph{Timestamp accuracy.}
Overall log-time discrepancy curves show that \emph{Interval} and \emph{Interval+Type} provide the most accurate timestamp predictions, whereas \emph{Zero-shot} exhibits the largest errors. When stratified by temporal distance from presentation, prompt differences are minimal for
immediate events but widen substantially at longer horizons (1~day, 1~week, 1~year, and beyond). Structured prompts with explicit temporal cues maintain accuracy across these
intervals, while unstructured prompts degrade more quickly.

When stratified by the time from presentation (e.g., within 1 hour, 1 day, 1 week), prompt differences are most pronounced for events occurring farther from the report anchor (Figure \ref{fig:prompt_ablation}, far right). For immediate events, all prompt versions perform comparably. However, at longer timescales (1 week to "ever"), Interval and Interval+Type maintain superior accuracy, while Zero-shot shows dramatic drops in performance. This suggests that structured prompts are particularly critical for accurate long-range temporal reasoning.

Across semantic similarity, temporal ordering, and timestamp alignment, prompt components that provide explicit temporal structure or event-typing requirements consistently improve
performance, while removing instructional scaffolding—particularly in the Zero-shot condition—results in substantial degradation. These findings highlight the importance of prompt
design in guiding LLMs to produce high-quality temporal annotations from clinical narratives.

\subsection{Inter-LLM Comparison of Temporal Annotations} 
\label{apd:inter-llm-comparison}

\begin{figure}[!htbp]
  \centering
  \begin{subfigure}[!htbp]{0.45\linewidth}\centering
    \includegraphics[width=0.9\linewidth]{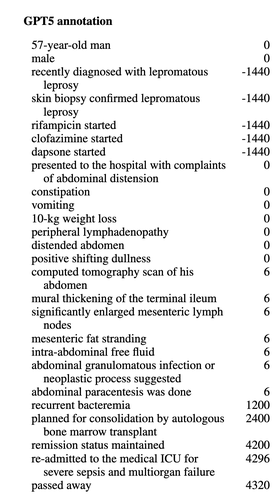}
  \end{subfigure}\hfill
  \begin{subfigure}[!htbp]{0.45\linewidth}\centering
    \includegraphics[width=0.9\linewidth]{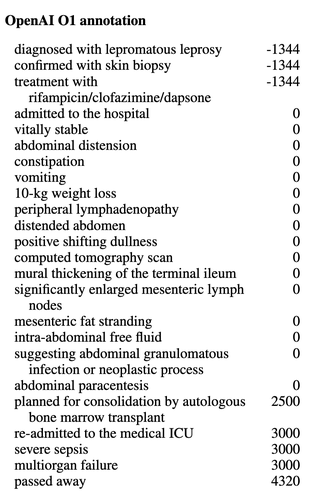}
  \end{subfigure}

  \vspace{2pt}

  \begin{subfigure}[!htbp]{0.45\linewidth}\centering
    \includegraphics[width=0.9\linewidth]{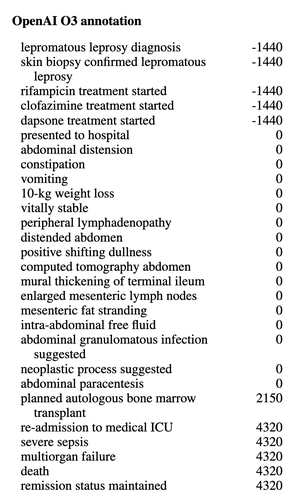}
  \end{subfigure}\hfill
  \begin{subfigure}[!htbp]{0.45\linewidth}\centering
    \includegraphics[width=0.9\linewidth]{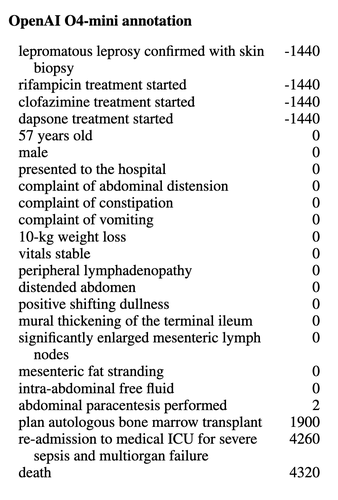}
  \end{subfigure}

\caption{Comparison of temporal annotations generated by several LLMs for a representative case report (PMC10629858). The figure contrasts manual annotations with model-generated event–time
sequences, highlighting differences in event granularity, phrasing, duplicate event generation, and temporal normalization (e.g., variation in converting narrative expressions such as ``two
months before admission’’ into hour offsets). Long-range timestamp assignments (e.g., for transplant planning, ICU readmission, and death) show the greatest variability across models, whereas the overall ordering of major clinical events is largely preserved. This example illustrates characteristic
differences in how LLMs interpret, segment, and temporally anchor clinical narratives.}
  \label{fig:compare_llm_annotations}
  \vspace{-3mm}
\end{figure}

To illustrate qualitative differences among annotators, we compare temporal event sequences generated by several LLMs for a representative case report (PMC10629858) \citep{abu2023rare}, which describes a patient with lepromatous leprosy who later developed abdominal complications, recurrent bacteremia, severe sepsis, and ultimately died (Figure~\ref{fig:compare_llm_annotations}).

\begin{figure}[!htbp]  %
  \centering
    \begin{tcolorbox}[colback=white,colframe=black!30,boxrule=0.6pt,left=2mm,right=2mm,top=2mm,bottom=2mm,width=\linewidth]
    \footnotesize \textbf{Excerpt from \cite{abu2023rare}:}
    \begin{quote}\footnotesize
    A 57-year-old man recently diagnosed with lepromatous leprosy was confirmed with skin biopsy and had been on treatment (rifampicin/clofazimine/dapsone) for 2 months before admission; he was presented to the hospital with complaints of abdominal distension, constipation, vomiting, and a 10-kg weight loss. On examination, the patient was vitally stable. He had evidence of peripheral lymphadenopathy with a distended abdomen and a positive shifting dullness. A computed tomography scan of his abdomen showed mural thickening of the terminal ileum with significantly enlarged mesenteric lymph nodes, mesenteric fat stranding, and intra-abdominal free fluid, suggesting abdominal granulomatous infection or neoplastic process.
    \dots 
    
    The patient was planned for consolidation by autologous bone marrow transplant. Unfortunately, with the recurrent bacteremia and sepsis that accompanied the patient’s course due to his immunocompromised state, he was re-admitted to the medical ICU for severe sepsis and multiorgan failure and passed away around 6 months after his initial diagnosis with NHL, despite maintaining a remission status.
    \end{quote}
    \end{tcolorbox}
  \begin{subfigure}[!]{0.45\linewidth}\centering
    \includegraphics[width=0.9\linewidth]{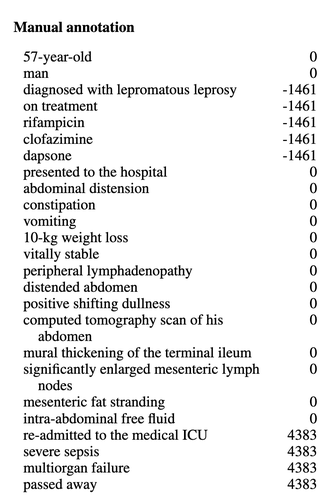}
  \end{subfigure}\hfill
  \begin{subfigure}[!htbp]{0.45\linewidth}\centering
    \includegraphics[width=0.9\linewidth]{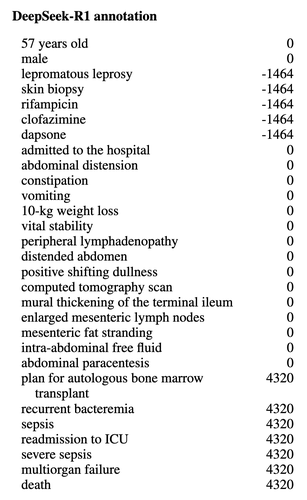}
  \end{subfigure}

  \begin{subfigure}[!htbp]{0.45\linewidth}\centering
    \includegraphics[width=0.9\linewidth]{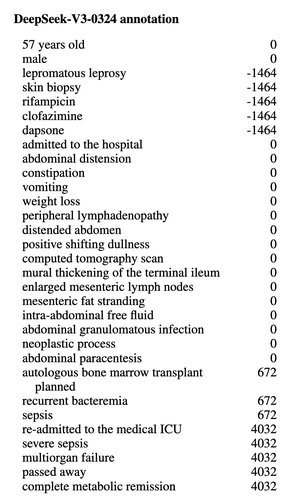}
  \end{subfigure}\hfill
  \begin{subfigure}[!htbp]{0.45\linewidth}\centering
    \includegraphics[width=0.9\linewidth]{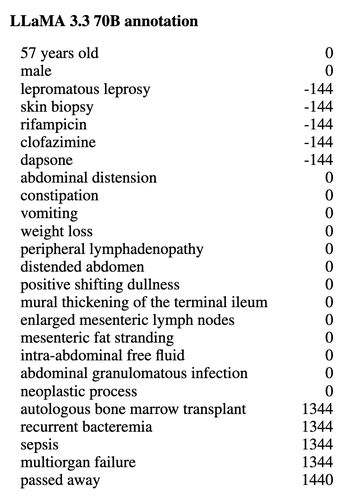}
  \end{subfigure}
\end{figure}

Across models, the principal clinical trajectory is consistently recovered: confirmation of lepromatous leprosy, antibiotic treatment, evolving abdominal symptoms with corresponding imaging findings, subsequent sepsis, ICU readmission, and death. All annotators preserve the
temporal polarity of events, assigning negative times to historical information and positive times to post-admission findings.

Differences emerge in event granularity and phrasing. Some models split conjunctive phrases into multiple events (e.g., listing each component medication separately), whereas others
compress them into single entries. Several annotators also produce near-duplicate events at the same timestamp (e.g., both “diagnosed with lepromatous leprosy’’ and “skin biopsy confirmed’’), reflecting variation in how models segment clinically related information within the narrative.

More pronounced differences appear in temporal normalization. For the expression “two months before admission,” models produce hour offsets ranging from approximately $-1464$ to $-1344$, and one model assigns $-144$ due to an apparent day–hour unit error. Long-range events such as the autologous bone marrow transplant plan, ICU readmission, and death also receive timestamps that vary by hundreds of hours across annotators. Despite this variability in absolute timing,
the relative ordering of major clinical milestones remains largely preserved.

A recurring issue across models involves distinguishing planned actions from events that actually occurred. For example, most annotators correctly represent “planned autologous bone-marrow transplant’’ with an associated future timestamp, which is appropriate because planned interventions can be temporally located and may influence subsequent decision-making. However, several models omit the word “planned,’’ converting the entry into “autologous
bone-marrow transplant’’ and thereby representing an intended but unperformed procedure as an event that occurred. In such cases, the discrepancy arises from the event label rather than the timestamp itself, illustrating how small linguistic omissions can alter the extracted temporal semantics.
\begin{table}[h!]
\centering
\caption{Common error types observed across LLM-generated temporal annotations.}
\label{tab:error_taxonomy}
\begin{tabular}{p{3.2cm} p{9cm}}
\toprule
\textbf{Error Type} & \textbf{Description and Examples} \\
\midrule

\textbf{Event granularity mismatch} 
& Models differ in splitting vs.\ compressing multi-component statements. 
For example, medications may appear as three separate events (``rifampicin'', ``clofazimine'', ``dapsone'') or as one combined entry. \\[6pt]

\textbf{Near-duplicate events} 
& Multiple paraphrased versions of the same clinical concept assigned to the same timestamp, such as 
``diagnosed with lepromatous leprosy’’ and 
``skin biopsy confirmed lepromatous leprosy.’’ \\[6pt]

\textbf{Temporal normalization variability} 
& Large differences in converting narrative expressions (e.g., ``two months before admission’’) to numerical timestamps, with offsets ranging from $-1464$ to $-1344$ hours and occasional scaling errors (e.g., $-144$ hours). Long-range events (e.g., transplant plan, ICU readmission, death) also vary by hundreds of hours across models. \\[6pt]

\textbf{Intent vs.\ occurrence confusion} 
& Some models omit modality markers such as ``planned,'' converting intended future events into events marked as completed. 
The timestamp itself may be reasonable; the error lies in the event label. \\[6pt]

\textbf{Phrasing and specificity variation} 
& Models differ in the level of detail used in event descriptions.  
Some outputs retain clinically precise phrases (e.g., ``mural thickening of the terminal ileum’’), while others produce generic events (e.g., ``abdominal abnormality’’). \\[6pt]

\textbf{Long-range timestamp degradation} 
& Accuracy decreases for events far from the admission anchor, with larger dispersion in timestamp predictions for events weeks or months after presentation. \\
\bottomrule
\end{tabular}
\vspace{-5mm}
\end{table}

Overall, the models capture the broad sequence of events but differ in how they segment conjunctive statements, handle near-duplicates, and resolve absolute times—particularly for
long-range events. A taxonomy of common annotation errors observed across models—including granularity differences, duplicate events, temporal normalization variability, and intent–occurrence misclassification—is summarized in Table~\ref{tab:error_taxonomy}.

\subsection{Sensitivity to Event Matching Strategy (LAPJV)}
\label{apd:lapjv-matching}
We additionally consider a globally optimal one-to-one event alignment based on the Linear Assignment Problem, solved using the Jonker--Volgenant (LAPJV) algorithm \citep{jonker1987shortest}. We define a rectangular cost matrix whose entries are pairwise cosine distances between predicted and reference events. To enforce the same distance thresholding used in the recursive heuristic, we assign a large penalty cost to pairs whose distance exceeds the threshold, effectively masking infeasible matches. LAPJV then computes a minimum-cost assignment under these constraints, yielding a globally optimal one-to-one alignment between the two event sets.

To assess sensitivity to the event alignment procedure, we repeat the inter-LLM comparison using LAPJV-based matching at a cosine distance threshold of 0.1. Table~\ref{tab:model_comparison_lapjv} reports event match rate, temporal concordance, and AULTC across five independent runs per model, and is directly comparable to the main-text results obtained using the recursive matching heuristic (Algorithm~\ref{alg:recursive_match}; Table~\ref{tab:model_comparison}).
\texttt{OpenAI GPT5} achieves the highest mean match rate (0.850 $\pm$ 0.006), indicating the most comprehensive coverage of manually annotated events. \texttt{OpenAI O3} attains the highest concordance (0.894 $\pm$ 0.004), and \texttt{OpenAI O4-mini} yields the highest AULTC (0.749 $\pm$ 0.003), reflecting strong timestamp alignment among matched events. Standard deviations are generally small across runs, suggesting stable behavior for a given model.

\begin{table}[h!]
\vspace{-3mm}
\centering
\caption{Inter-LLM comparison under LAPJV-based event matching (threshold = 0.1), evaluated on 200 manually annotated case reports across five runs per model.}
\label{tab:model_comparison_lapjv}
\begin{tabular}{lccc}
\hline
\textbf{Model} & \textbf{Match Rate} & \textbf{Concordance} & \textbf{AULTC} \\
\hline
\texttt{Llama 3.3 70B}       & 0.678 $\pm$ 0.003 & 0.838 $\pm$ 0.003 & 0.717 $\pm$ 0.001 \\
\texttt{DeepSeek-R1}         & 0.749 $\pm$ 0.009 & 0.790 $\pm$ 0.009 & 0.726 $\pm$ 0.004 \\
\texttt{DeepSeek-V3-0323}    & 0.764 $\pm$ 0.005 & 0.793 $\pm$ 0.007 & 0.728 $\pm$ 0.002 \\
\texttt{OpenAI GPT5}        & \textbf{0.850 $\pm$ 0.006} & 0.883 $\pm$ 0.003 & 0.719 $\pm$ 0.001 \\
\texttt{OpenAI O4-mini}      & 0.752 $\pm$ 0.021 & 0.843 $\pm$ 0.008 & \textbf{0.749 $\pm$ 0.003} \\
\texttt{OpenAI O3}           & 0.739 $\pm$ 0.004 & \textbf{0.894 $\pm$ 0.004} & 0.736 $\pm$ 0.003 \\
\texttt{OpenAI O1}           & 0.640 $\pm$ 0.008 & 0.881 $\pm$ 0.004 & 0.731 $\pm$ 0.001 \\
\hline
\end{tabular}
\vspace{-5mm}
\end{table}

\begin{figure}[h!]
    \centering
    \includegraphics[width=\linewidth]{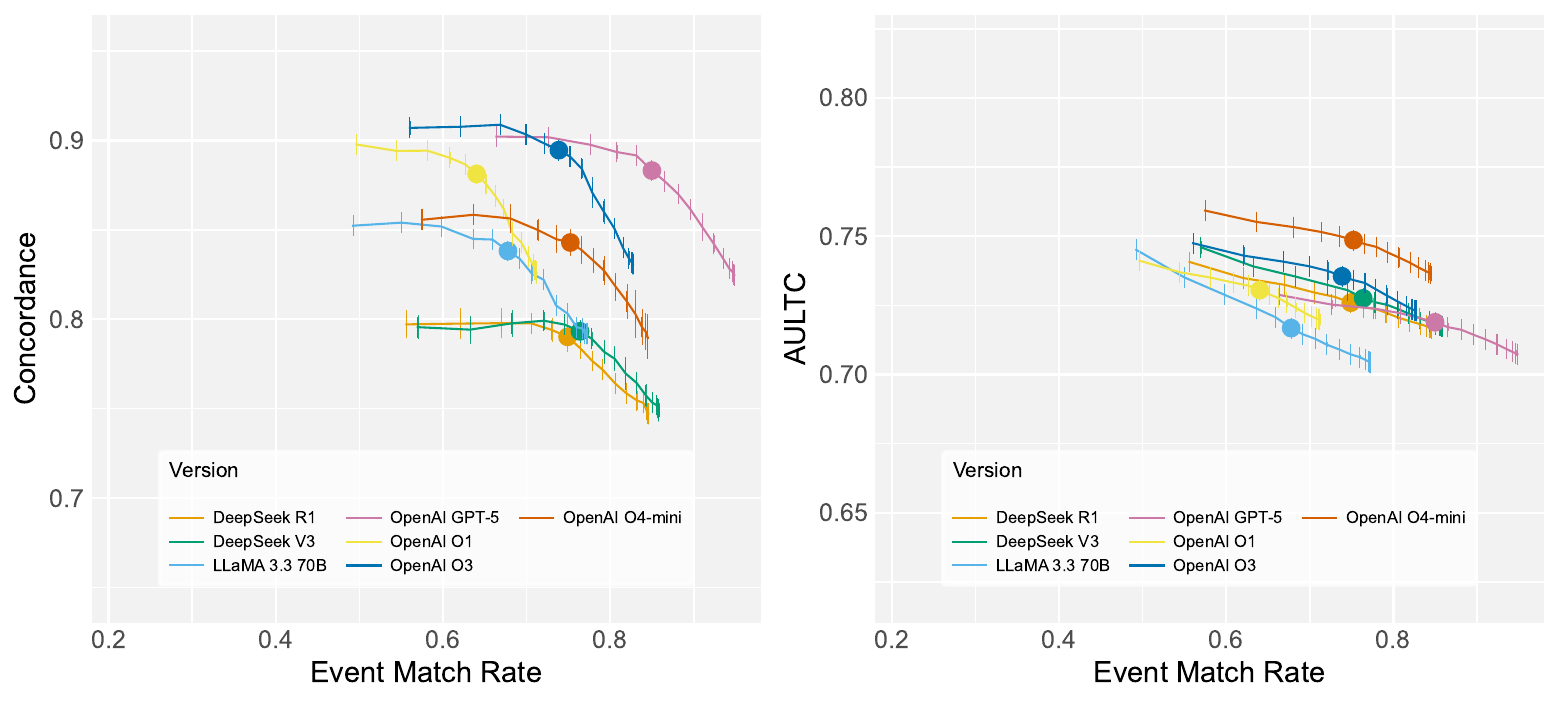}
    \caption{\textbf{A:} Temporal concordance versus event match rate under LAPJV-based matching.
    \textbf{B:} AULTC versus event match rate. Solid circles ($\bullet$) indicate the operating
    threshold of 0.1; ticks (|) denote 0.01 threshold increments in [0.01, 0.50].}
    \label{fig:threshold_tradeoff_LAPJV}
    \vspace{-5mm}
\end{figure}

We further examine the trade-off between event match rate and temporal accuracy by sweeping
the LAPJV distance threshold from 0.01 to 0.50 in increments of 0.01. 
Figure~\ref{fig:threshold_tradeoff_LAPJV}
plots temporal concordance and AULTC as a function of event match rate, with the operating
point at threshold 0.1 highlighted.

Across models, LAPJV preserves high event match rates at low thresholds and exhibits a
smooth trade-off between coverage and temporal accuracy. 
The resulting operating points in Figure~\ref{fig:threshold_tradeoff_LAPJV}
are highly comparable to those obtained using the recursive matching heuristic in Figure~\ref{fig:threshold_tradeoff_greedy} (e.g., at $\tau=0.1$, the operating point used in our main results in Table~\ref{tab:model_comparison}). 
In particular,
the heuristic matcher yields slightly stronger concordance near the elbow of the trade-off
curve, while LAPJV achieves marginally higher AULTC values, reflecting improved timestamp
alignment under global assignment. 
Overall, both strategies lead to consistent qualitative
conclusions.

\section{Survival Analysis Pipeline}
\label{apd:survival-analysis-pipeline}

We developed a survival analysis framework to evaluate the prognostic value of textual information encoded in the time series for time until death prediction. 
Our approach follows a two-stage process:

\begin{enumerate}
\item \textbf{Embedding Extraction}: Each textual time series is converted to a textual context in the format \texttt{"(time) clinical event [SEP] ... [SEP] (time) clinical event [SEP]"} and processed using various pre-trained language models to obtain dense vector representations. 
For encoder models (BERT, RoBERTa, ModernBERT, etc.), we extract the \texttt{[CLS]} token embedding, while for decoder models (\texttt{Llama}, \texttt{DeepSeek-R1}), we compute the mean-pooled embedding over all non-padding tokens.

\item \textbf{Survival Modeling}: These embeddings serve as covariates for two survival models:
\begin{itemize}
    \item \textbf{DeepSurv} \citep{katzman2018deepsurv}: A neural network generalization of the Cox proportional hazards model.
    \item \textbf{DeepHit} \citep{lee2018deephit}: A multi-task neural network jointly modeling discrete hazard and survival functions.
\end{itemize}
\end{enumerate}

We evaluated the performance of the survival models using two metrics: the time-dependent concordance index \citep{antolini2005time}, which measures the model’s ability to correctly rank survival times, and the integrated Brier score (IBS), which captures overall calibration and discrimination over time. 

To ensure reliable estimates, we conducted experiments across five random seeds.
For each seed, we performed a train-validation-test split with proportions of 64\%, 16\%, and 20\%, respectively. 

All survival times were right-censored at a fixed cutoff (\texttt{max\_right\_censoring\_time}=1 year) to maintain consistency across samples. 
Only samples with positive survival times after censoring were included for model training and evaluation.

We tuned hyperparameters independently for each combination of language model and survival model. 
For DeepSurv and DeepHit, we searched over the following hyperparameter grid:
\begin{itemize}
    \item \texttt{num\_nodes}: \{64, 128, 256, 512, 1024, 2048, 4096\}
    \item \texttt{dropout}: \{0.1, 0.5\}
    \item \texttt{epochs}: 2000 (with early stopping)
\end{itemize}

For each configuration, the model was trained on the training set and evaluated on the validation set using the time-dependent concordance index. 
The configuration yielding the highest validation time-dependent concordance was selected, and the corresponding model was then evaluated on the held-out test set. 
This process was repeated for each random seed.
We report the mean and standard deviation of the concordance-index and IBS across the five seeds for each combination of language model and survival model. 

\paragraph{Computational resources:}
For embedding generation in the survival modeling task, decoder models (\texttt{Llama}, \texttt{DeepSeek-R1}) used mean-pooled hidden states, and encoder models (e.g., PubMedBERT) used \texttt{[CLS]} embeddings. Survival models (DeepSurv, DeepHit) were trained using PyTorch on single NVIDIA GeForce RTX 4090. Each training run optimizing across the hyperparameter grid for the validation set took on average 195 minutes. Hyperparameter search across seeds, dropout values, and hidden node counts resulted in 48 GPU-hours total compute.

\end{document}